\pdfoutput=1
\documentclass[11pt]{article}

\usepackage{EMNLP2022}

\usepackage{times}
\usepackage{latexsym}

\usepackage[T1]{fontenc}
\usepackage[utf8]{inputenc}

\usepackage{microtype}
\usepackage{url}
\usepackage{times}
\usepackage{latexsym}
\usepackage{graphicx}
\usepackage{todonotes}
\usepackage{subcaption}
\usepackage{amsmath}
\usepackage{booktabs}
\usepackage{multirow}
\usepackage{todonotes}
\usepackage{wasysym}
\usepackage{tabularx}
\usepackage{pgfplots}
\usepackage{amssymb}% http://ctan.org/pkg/amssymb
\usepackage{pifont}% http://ctan.org/pkg/pifont
\newcommand{\cmark}{\ding{51}}%
\newcommand{\xmark}{\ding{55}}%

\usepackage{enumitem}
\setlist[itemize]{align=parleft,left=0pt..1em}

\newcommand{\MA}{$\mathcal{M}$}
\newcommand{\MG}{$\mathcal{M}_G$}
\newcommand{\WS}{\textbf{WS}}
\newcommand{\LW}{\textbf{LW}}
\newcommand{\SCI}{\textbf{SCI}}
\newcommand{\SCO}{\textbf{SCO}}
\newcommand{\SC}{\textbf{SC}}

\newcommand{\YES}{\textbf{`[\_YES\_]'}}
\newcommand{\PROBABLY}{\textbf{`[\_PROBABLY\_]'}}
\newcommand{\MAYBE}{\textbf{`[\_MAYBE\_]'}}
\newcommand{\DOUBT}{\textbf{`[\_DOUBT\_]'}}
\newcommand{\NO}{\textbf{`[\_NO\_]'}}

\newcommand{\mypara}[1]{\noindent\textbf{#1}}

\title{Knowledge Transfer from Answer Ranking to Answer Generation}

\author{Matteo Gabburo$^{1}$\thanks{\ \ Work done as an intern at Amazon Alexa AI}\ , Rik Koncel-Kedziorski$^{2}$, Siddhant Garg$^{2}$, \\ \textbf{Luca Soldaini$^{3}$\thanks{\ \ Work completed at Amazon Alexa AI}\ , Alessandro Moschitti$^{2}$}\\
$^{1}$University of Trento , $^{2}$Amazon Alexa AI, $^{3}$Allen Institute for AI\\
\texttt{matteo.gabburo@unitn.it} \\ \texttt{\{rikdz,sidgarg,amosch\}@amazon.com} \\
\texttt{lucas@allenai.org} \\
}

\begin{document}
\maketitle
\begin{abstract}
Recent studies show that Question Answering (QA) based on Answer Sentence Selection (AS2) can be improved by generating an improved answer from the top-$k$ ranked answer sentences (termed GenQA). This allows for synthesizing the information from multiple candidates into a concise, natural-sounding answer. However, creating large-scale supervised training data for GenQA models is very challenging. In this paper, we propose to train a GenQA model by transferring  knowledge from a trained AS2 model, to overcome the aforementioned issue. First, we use an AS2 model to produce a ranking over answer candidates for a set of questions. Then, we use the top ranked candidate as the generation target, and the next $k$ top ranked candidates as context for training a GenQA model. We also propose to use the AS2 model prediction scores for loss weighting and score-conditioned input/output shaping, to aid the knowledge transfer. Our evaluation on three public and one large industrial datasets demonstrates the superiority of our approach over the AS2 baseline, and GenQA trained using supervised data.

\end{abstract}

\vspace{-.5em}
\section{Introduction}
\vspace{-.2em}
\label{sec:introduction}
In recent times, extractive QA research can be categorized into two broad directions for the task of producing the final answer for a question: (i) Answer Sentence Selection (AS2), which, given a question and a set of answer-sentence candidates, selects sentences (e.g., retrieved by a search engine) that correctly answer the question; and (ii) Machine Reading (MR), e.g., \cite{Chen-Fisch-2017}, which, given a question and a reference text, involves finding an exact text span that answers the question.
AS2 models can perform more efficiently with large text databases (as they originated from the TREC-QA track \cite{voorhees99trec}), and there seems a renewed research interest in these models for applications to personal assistants, e.g., Alexa \cite{garg2020tanda,DBLP:conf/sigir/MatsubaraVM20,garg-moschitti-2021-will}.  

Both approaches (AS2 and MR) when applied for QA over unstructured web text, while effective, may have certain drawbacks. Arbitrary web sentences may not contain all the information needed to answer a question, or may contain distracting extraneous information. Moreover, they may have a particular sentiment or style that is not suited to QA, or be too structurally reliant on longer discourse context to serve as a standalone answer. 
In light of this, researchers have been exploring text generation systems for writing `better' answers. For example, in MR, RAG~\cite{10.5555/3495724.3496517} generates an answer from a set of documents selected by dense passage retrieval models.

For AS2 systems, research has focused on learning to summarize answers from relevant paragraphs~\cite{lewis-etal-2020-bart}, or to synthesize information from the top ranked candidates of an AS2 system~\cite{hsu2021answer}. The latter approach, termed as GenQA, has shown improvements in terms of both answer accuracy and style suitability. 
A distinctive characteristic of GenQA over a generation-based approach for MR is the length of the answer: the former uses an entire sentence as the target, while the latter in practice uses a short text (primarily targeting entity names). In this work, we focus on GenQA as we are interested to generate complete answer sentences from precise information selected by AS2 models.

A challenge for training effective GenQA models is the difficulty of obtaining large-scale, high-quality training data. Producing such data for GenQA typically requires human annotators to read questions and paragraphs of relevant background information, and then author a self-contained, natural answer (typically a sentence). This fairly involved procedure highly diminishes the velocity of annotation. Existing datasets in research works either offer limited coverage of all domains, where GenQA can be applied~\cite{bajaj2018ms}, or are too small to be used as supervised training data~\cite{DBLP:journals/corr/abs-2110-07150}. Generally, collecting a human-authored answer to a question when given a context is significantly more expensive compared to annotating the correctness of an extracted web sentence as an answer for the same question. Consequently, there are a large number of annotated datasets~\cite{wang2007jeopardy,yang2015wikiqa,garg2020tanda} available for the latter type, aimed at training answer sentence selection (AS2) systems.
 
In this work, we propose a training paradigm for transferring the knowledge learned by a discriminative AS2 ranking model to train an answer generation QA system. Towards this, we learn a GenQA model using weak supervision provided by a trained AS2 model on a unlabeled data set comprising of questions and answer candidates. Specifically, for each question, the AS2 model is used to rank a set of answer candidates without having any label of correctness/incorrectness for answering the question. The top ranked answer is used as the generation target for the GenQA model, while the question along with the next $k$ top-ranked answers are used as the input for the GenQA model.

We supplement the ranking order of answer candidates with the prediction confidence scores provided by the AS2 model for each answer candidate. This is done by modifying our knowledge transfer strategy in two ways. First, we weight the loss of each training instance (question + context, comprised of $k$ answer candidates) using the AS2 model score of the top ranked answer, which is to be used as the GenQA target. This allows the GenQA model to selectively learn more from `good' quality target answers in the weakly supervised training data (AS2 models are calibrated to produce higher confidence scores for correct answers). However, this loss weighting only considers the score of the output target, and does not exploit the scores of the input candidates. To overcome this limitation, we discretize and label the AS2 scores into $l$ confidence buckets, add these bucket labels to the GenQA vocabulary, and finally prepend the corresponding label to each answer candidate in the input and/or the output. This confidence bucket label provides the GenQA model with an additional signal about the answer quality of each candidate as assigned by the AS2 model. We show that both these techniques improve the QA accuracy, and can be combined to provide additional improvements.

We empirically evaluate~\footnote{We will release code and all trained models checkpoints at \url{https://github.com/amazon-research/
wqa-genqa-knowledge-transfer}} our proposed knowledge transferring technique from AS2 to GenQA on three popular public datasets: MS-MARCO NLG~\cite{bajaj2018ms}, WikiQA~\cite{yang2015wikiqa}, TREC-QA~\cite{wang2007jeopardy}; and one large scale industrial QA dataset. Our results show that the GenQA model trained using our paradigm of weak supervision from an AS2 model can surprisingly outperform both the AS2 model that was used for knowledge transfer (teacher), as well as a GenQA model trained on fully supervised data. On small datasets such as WikiQA and TREC-QA, we show that AS2 models trained even on small amounts of labeled data can be effectively used to weakly supervise a GenQA model, which then can outperform its teacher in QA accuracy. Additionally, on MS-MARCO NLG, where fully supervised GenQA training data is available, we show that an initial round of training with our weakly supervised methods yields additional performance improvements compared to the standard supervised training of GenQA. Qualitatively, the answers generated by our model are often more directly related to the question being asked, and stylistically more natural-sounding and suitable as responses than answers from AS2 models, despite being trained only on extracted sentences from the web.

\vspace{-.5em}
\section{Related Work}
\vspace{-.2em}
\label{sec:relatedwork}
Our work builds upon recent research in AS2, answer generation for QA, and transfer learning. 

\paragraph{Answer Sentence Selection} Early approaches for AS2  use CNNs~\cite{Severyn2015LearningTR} or alignment networks~\cite{Shen2017InterWeightedAN,Tran2018TheCA,Tay2018MultiCastAN} to learn and score question and answer representations. Compare-and-aggregate architectures have also been extensively studied~\cite{Wang2017ACM,Bian2017ACM,Yoon2019ACM} for AS2. ~\citet{Madabushi2018IntegratingQC} exploited fine-grained question classification to further improve answer selection. ~\citet{garg2020tanda} achieved state-of-the-art results by fine-tuning transformer-based models on a large-scale QA dataset first, and then adapting to smaller AS2 datasets.~\citet{Matsubara2020RerankingFE} combine multiple heterogeneous systems for AS2 to improve a QA pipeline, similar in spirit to GenQA. Several follow-up works have further improved the performance of AS2 using transformer models, using multiple answer candidates~\cite{zhang-etal-2021-joint} and document-aware pre-training strategies~\cite{di-liello-etal-2022-paragraph,diliello2021pretraininingtransformersas2}.

% \todo{Are we sure there is nothing else the relwork is small. I differentiate MR from GenQA }  
\paragraph{Answer Generation for QA} Answer generation for MR has been studied by ~\citet{Izacard2021LeveragingPR,10.5555/3495724.3496517}, while ~\citet{Iida2019ExploitingBK,Goodwin2020TowardsZS,Deng2020JointLO} have studied question-based summarization (QS). ~\citet{DBLP:journals/corr/abs-2112-08688} incorporate the evidentiality of retrieved passages for training a generator, evaluated for the QA task of open-domain MR span-extraction. ~\citet{DBLP:journals/corr/abs-2110-06393} obtain extractive answer spans from a generative model by leveraging the decoder cross-attention patterns. ~\citet{fajcik-etal-2021-r2-d2} combine a generative reader with an extractive reader to aggregate evidence from multiple passages for open-domain span-extraction.

\begin{figure}[t]
\includegraphics[width=\linewidth]{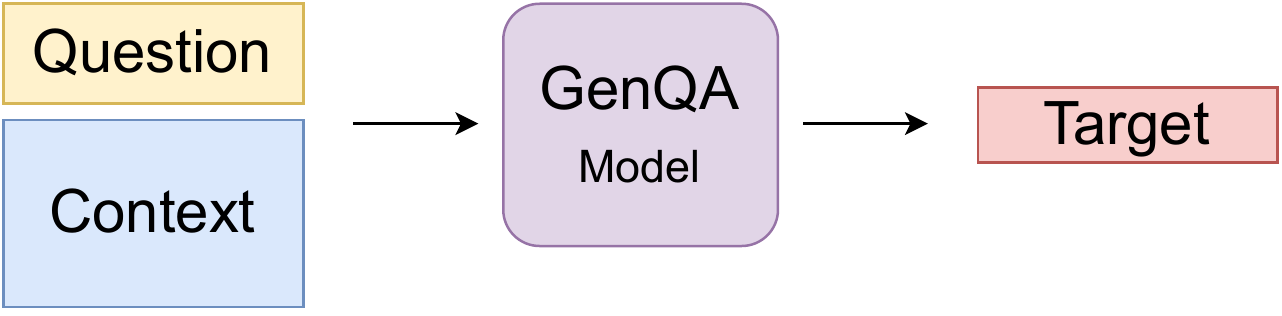}
\centering
\vspace{-2em}
\caption{\small A GenQA model \cite{hsu2021answer} is a seq2seq model that takes in input a question and $k$ answer candidates, and generates an answer.}
\label{fig:genqa}
\vspace{-.2em}
\end{figure}

All the previously described approaches focus on identifying short answer spans for answering questions. Research on generating complete sentences as answers (similar to answer sentences produced by extractive AS2 systems) is rarer, but includes~\citet{hsu2021answer}, that propose a QA pipeline for GenQA (refer Fig~\ref{fig:genqa}). This pipeline starts with an AS2 model that selects `good' answer candidates that are then used for generating the answer. ~\citeauthor{hsu2021answer} learn to generate natural responses to questions using the top ranked candidates from the AS2 model as input context to the GenQA model. GenQA has also been explored for multilingual QA ~\cite{DBLP:journals/corr/abs-2110-07150} by extending the answer generation approach to a multilingual setting, where the answer candidates (that are used as input to the GenQA model) can be from a mix of different languages. 

In all  these works, a major challenge is finding training data for effectively training GenQA models, which requires annotator-authored natural responses. In this work, we alleviate this problem by showing that it is possible to use AS2 ranked candidates to create the input context and output target for training GenQA, achieving state-of-the-art results.

\vspace{-.5em}
\paragraph{Transfer Learning} Transfer learning is well studied in NLP, including pre-training~\cite{devlinetal2019,liu2019roberta}, multi-task learning~\cite{luong2015multi}, cross-lingual transfer~\cite{Schuster2019CrosslingualTL} and domain adaptation~\cite{Gururangan2020DontSP}. Our work is squarely located in this space: our underlying language models are based on pre-training for text generation~\cite{radfordetall2019gpt2,raffeletall2020}; our main contribution is to show that knowledge can be transferred sequentially from a ranking (discriminative) task to a generation task. Recently \citet{wang2021gpl} propose a new domain adaptation method leveraging large unlabeled datasets and a query generator model. \citeauthor{Izacard2021LeveragingPR} used retrieved text passages containing evidences to train a generative model for open domain QA.

\vspace{-.5em}
\section{Knowledge Transfer: AS2 ${\rightarrow}$ GenQA}
\begin{figure}[t]
\includegraphics[width=\linewidth]{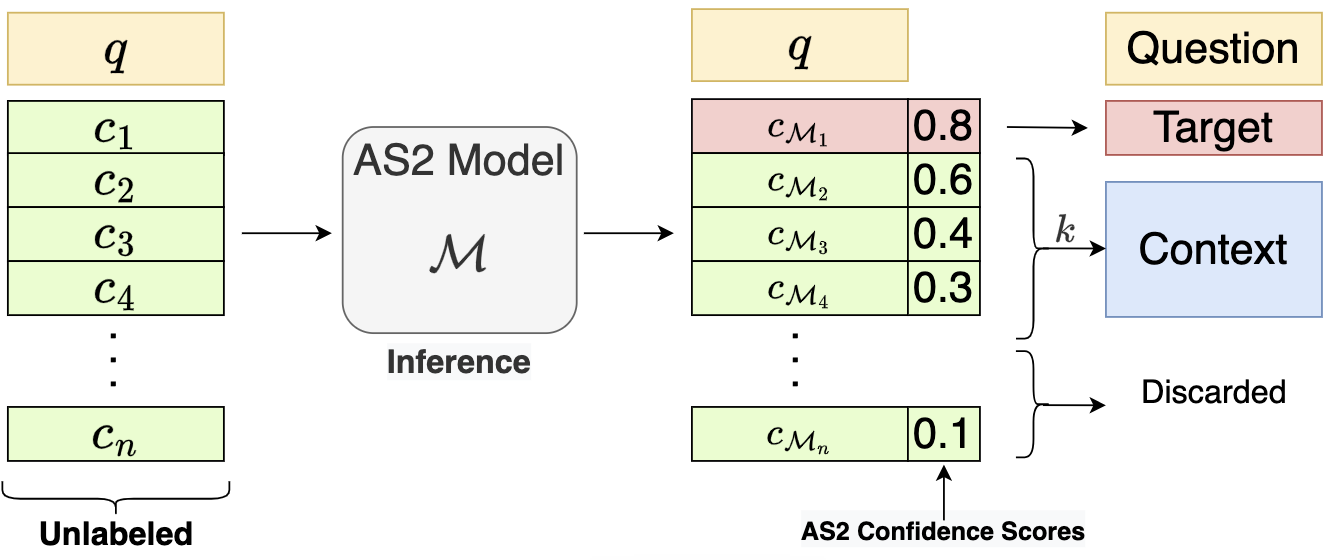}
\centering
\vspace{-2em}
\caption{\small Our pipeline for creating weakly supervised training examples for training GenQA models. The AS2 model assigns a confidence score to each answer candidate sentence. These scores are used to select the inputs and the target sequences for the GenQA model.}
\label{fig:weak_supervision}
\vspace{-.5em}
\end{figure}

\vspace{-.2em}
\label{sec:genqa_from_as2}
Previous works on GenQA require the use of labeled data for effectively training the GenQA model.
To reduce the need of expensive large-scale training data for GenQA, we propose a training paradigm that uses unlabeled data while being weakly-supervised by a discriminative AS2 model (as shown in Fig. \ref{fig:weak_supervision}).

\vspace{-.4em}
\subsection{Answer Sentence Selection (AS2)}
\vspace{-.2em}
AS2 is a popular task in QA, defined as follows: Given a question $q$, and a set of answer candidates $C=\{c_1,\dots,c_n\}$ (retrieved using a web-index, KB, etc), find the answer candidate $c_q \in C$ that best answers $q$. This is typically modeled as a binary classifier {\MA} over QA pairs, labeled as correct or incorrect. At inference, the scores assigned by {\MA} can be used to produce a ranking over $C$, with $c_q= \text{argmax}_{i} \ \mathcal{M}(q,c_i) $.

\vspace{-.2em}
\subsection{Generative QA (GenQA)}
\vspace{-.2em}
Generative QA refers to using a text generation model for generating an answer for a question. More specifically, when provided with a question $q$ and context $\bar{c}$, the GenQA model {\MG} should generate a natural sounding answer $c_q=\mathcal{M}_G(q,\bar{c})$ that correctly answers $q$. 
Following \citet{hsu2021answer}, we consider a set of $k$ answer candidates as the context $\bar{c}$ to be provided to {\MG}. 

\vspace{-.2em}
\subsection{Training GenQA using an AS2 model}
\vspace{-.2em}
\label{ssec:training_genqa}

We aim at training a GenQA model, {\MG}, using a trained AS2 model, {\MA}, which predicts correctness/incorrectness of  answer candidates for a given question. Specifically, we use an unsupervised dataset, $\mathcal{U}$, comprising of a set of questions along with their retrieved answer candidates, i.e, $(q,C=\{c_1,\dots,c_n\})$. Note that there are \emph{no} human annotations of correctness/incorrectness for the answer candidates in $C$ for the question $q$.

For each question $q \in \mathcal{U}$, we denote the ranking of answer candidates by {\MA} in decreasing order of prediction scores by $C_{M}=\{c_{M_1},c_{M_2},\dots,c_{M_n}\}$. We create weakly supervised examples for training the GenQA model by using
$\big (q$, $\bar{c}=\{c_{M_2},c_{M_3},\dots, c_{M_{k+1}}\}\big)$ as the input, and setting the generation target to be the top ranked answer candidate from {\MA}, i.e, $c_{M_1}$. For seq2seq transformer-based text generation models such as T5~\cite{raffeletall2020} and BART~\cite{lewis-etal-2020-bart}, we concatenate the question and $k$ answer candidates: ``$q$ [SEP] $c_{M_2}$ [SEP] $\dots$ [SEP] $c_{M_{k+1}}$''  to be provided as input to {\MG} and use the negative log probability of predicting each token of the target $c_{M_1}$ given the previous tokens as the training loss.

The resulting GenQA model {\MG} is trained on the unsupervised dataset only using \emph{weak supervision} from the discriminative AS2 model {\MA}. For the rest of the paper, we denote this training paradigm for GenQA by {\WS}. This approach is related to knowledge distillation (KD)~\cite{hinton2015distilling} wherein the predictions of a teacher model are used for guiding the learning of a student model. The novelty of our proposed approach from standard distillation techniques stems from the fact that the teacher (AS2) and student (GenQA) belong to different paradigms of training, the former being a discriminative classifier model while the latter being a generative model. Furthermore, standard KD techniques~\cite{hinton2015distilling,sanh2020distilbert} use a combination of supervision from the teacher (KL divergence) and supervision from the labeled data (Cross Entropy) for teaching the student, while in our case, we only use the supervision signal from the teacher without any access to labeled data.

\vspace{-.4em}
\subsection{Weighting GenQA Loss with AS2 scores}
\vspace{-.2em}
\label{ssec:lossweighting}

The binary cross-entropy loss used for training discriminative AS2 models typically calibrates their prediction w.r.t answer correctness~\cite{kamath-etal-2020-selective,garg-moschitti-2021-will}. This means that the top ranked answer to a question from {\MA} that receives a high prediction probability is more likely to be correct than the answer to another question that receives a lower prediction probability. We exploit this in addition to the ranking order generated by {\MA} to improve the learning of the GenQA model {\MG}. Intuitively, we want the GenQA model to learn more from `good' quality target answers (having higher prediction scores) than from lower quality answers.

To this end, we propose to modify our {\WS} cross-entropy loss by incorporating the AS2 scores provided by the AS2 model {\MA} when performing the knowledge transfer. Specifically, we use the prediction score $\mathcal{M}(q,c_{M_1})$ (normalized in $[0,1]$) of {\MA} on the top ranked answer candidate $c_{M_1}$ to weight the loss term for {\MG} corresponding to that instance (question $q$). Formally, the loss for each last generated word, $y_r$ of the generated output $y$ is:
\vspace{-1.5em}\\
\begin{equation}
\label{eq:lw}
\hspace{-.2em}
    \mathcal{L}_{\mathcal{M}_G}(q, c_{M_1}) = \frac{1}{Z}\mathcal{M}(q,c_{M_1}) \times \mathcal{L}_{G}(y_r,c_{M_1}), \hspace{-.5em}
\vspace{-.5em}
\end{equation}
%
%-\log \mathcal{P}(c_{M_1}|q,c_{M_2} \ldots c_{M_k}),
where $Z$ is the normalizing constant for AS2 scores computed on the training dataset, $\mathcal{L}_{G}$ is the standard loss for generating $y_r$, and $c_{M_1}$ is assumed to be the gold standard output.
$\mathcal{L}_{G}$ is defined as:
\vspace{-1.5em}\\
\begin{displaymath}
    \mathcal{L}_{G}(y_r, c_{M_{1}})=-\sum_{v\in V}log\frac{e^{y_r(v)}}{\sum_{h \in V} e^{y_r(h)}}c_{M_{1}}(r,v),
\vspace{-.5em}
\end{displaymath}
where $V$ is the vocabulary, $y_r(v)$ is the score of generating the word $v$ at position $r$, and $c_{M_{1}}(r,v)$ is 1 if the $r^{th}$ word of $c_{M_{1}}$ is $v$, otherwise it is -1.

We refer to the model trained with Eq.~\ref{eq:lw} as {\LW}.

\vspace{-.4em}
\subsection{AS2 Score Conditioned I/O Shaping}
\vspace{-.2em}
\label{ssec:SIOLF}

In the previous section, we described how to use $\mathcal{M}(q,c_{M_1})$ -- the AS2 prediction score of $c_{M_1}$ -- to weight the training loss for question $q$, since this candidate is used as the target for $q$ in {\MG}. However {\LW} ignores the AS2 scores for the other answer candidates $c_{M_2}\ldots c_{M_k+1}$, and does not explicitly provide this AS2 score as context to the GenQA model. To overcome this, we propose a method for labeling each candidate in the input of {\MG} with a representation of its AS2 score. This method can also be applied to the model output, which results in an improved performance (as shown in Section~\ref{sec:experiments}).

We define a bucketing function $\mathcal{F}$ over the normalized interval $[0,1]$ that operates on the AS2 prediction score $\mathcal{M}(q,a)$. For a QA pair $(q,a)$, $\mathcal{F}(q,a)$ assigns a confidence bucket label $b_i \in [b_1,\dots,b_l]$ based on $\mathcal{M}(q,a)$. Here $\mathcal{F}(q,a)$ is assigned $b_i$ if $\mathcal{M}(q,a)$ is in the interval $\Big[\frac{i-1}{l},\frac{i}{l}\Big)$. For our experiments, we set the value of $l{=}5$. We add the bucket labels $b_i$ as special tokens to the vocabulary of \MG.\footnote{We experimented with using existing tokens from the vocabulary as bucket labels (e.g. ``Probably'',``Maybe'') in hopes of reusing model knowledge about the semantics of these words, but obtained worse empirical results.}

We use $\mathcal{F}$ to modify the input and output of the GenQA model as follows:
\vspace{-.5em}
\begin{itemize}
    \item \textbf{AS2 Score Conditioned Input ({\SCI}):} We prepend the bucket label $b_j = \mathcal{F}(q,c_{M_j})$ to each of the $j \in \{2,{\dots},{k{+}1}\}$ answer candidates to be provided as input to {\MG}, so that the new input is formatted as: ``$q$ [SEP] $b_2 \ c_{M_2}$ [SEP] $\dots$ [SEP] $b_{k+1} \ c_{M_{k+1}}$''.
    \vspace{-.5em}
    \item \textbf{AS2 Score Conditioned Output ({\SCO}):} We prepend the bucket label $b_1 = \mathcal{F}(q,c_{M_1})$ to the target answer candidate $c_{M_1}$, so that the output target of {\MG} is: ``$b_1 \ c_{M_1}$''. 
\end{itemize}
\vspace{-.5em}
\noindent {\SCI} and {\SCO} can be used independently as well as jointly for training the GenQA model {\MG} using {\MA}. For simplicity, we will use the acronym {\SC} when these two techniques are used together. 

We propose {\SCI} to make the knowledge transfer more effective. Intuitively, labeling each input candidate with a special token correlated with its AS2 score helps the GenQA model: during training the model can focus more on the answer candidates associated with higher quality (more correct answers), thereby improving the model performance.

While {\SCO} is related to {\LW} presented in Sec~\ref{ssec:lossweighting}, it differs in the fact that the former allows the model to ``know'' the score of the target when designing internal representations of the text in its input and output. We hypothesize that this knowledge allows the model to organize its internal representations differently in the presence of bad targets, rather than just be less influenced by them as in {\LW}. Another advantage of {\SCO} is that during inference time, we can use the generated bucket label token as a confidence score for the GenQA model's answer. Calibrating confidence scores for text generation models, e.g., using sequence likelihood, etc. is challenging, especially when decoding is constrained as in real world applications. Finally, we can force decoding to start from any one of the {\SCO} bucket tokens in order to exploit its influence on the model's output. We empirically explore this in Appendix~\ref{apx:quality_eval}.

\vspace{-.5em}
\section{Datasets and Models}
\vspace{-.2em}
\label{sec:data}
For training and evaluating our knowledge transfer techniques ({\WS}, {\LW}, {\SC}) described above, we categorize the data that we use for each experiment into the following four sources/types:
\vspace{-.5em}
\begin{itemize}
    \item\textbf{AS2:} Labeled $(q,a)$ pairs with correctness and incorrectness annotation for training {\MA}
    \vspace{-.5em}
\item\textbf{Transfer:} Unlabeled $(q,a)$ pairs that are ranked by {\MA}, and used for knowledge transfer to {\MG}
    \vspace{-.5em}
\item\textbf{Fine-tuning:} Labeled data (human written answers / answers with correctness labels) for fine-tuning {\MG}, \emph{whenever available} 
\vspace{-.5em}
    \item\textbf{Evaluation:} Evaluation data for {\MA} and {\MG}
\end{itemize} 
\vspace{-.5em}

\noindent In Section~\ref{sec:experiments}, we vary the sources of different types of the data described above, to demonstrate the robustness and generality of our knowledge transfer method. Below, we provide details about the data sources we use, along with a summary of the underlying models. 

\vspace{-.4em}
\subsection{Unlabeled Data}
\vspace{-.2em}
\label{ssec:unlabeled_data}

\mypara{MS-MARCO QA} A popular MR dataset released by ~\cite{bajaj2018ms}. We use the training split which contains ${\sim}$800k unique user queries from the Bing search engine along with $\sim10$ passages retrieved for each question.\footnote{MS-MARCO v2.1:  \url{https://huggingface.co/datasets/ms_marco}} We split the original dataset into individual sentences using the BlingFire tokenizer\footnote{\url{https://github.com/microsoft/BlingFire}} to be used as the {\bf Transfer} data. Note that this dataset is used as unlabeled data for our experiments.

\mypara{AQAD-U} A large scale internal industrial QA dataset containing \emph{non-representative de-identified} user questions from Alexa virtual assistant. This unlabeled Alexa QA Dataset (AQAD-U) contains ${\sim}$50 million questions, and ${\sim}$400 answer candidates retrieved for each question using a large scale web index that contains over 100M web documents. We use this dataset as {\bf Transfer} data for experiments in the industrial setting.

\vspace{-.5em}\subsection{Labeled Data}
\label{ssec:labeled_data}
\vspace{-.2em}

\mypara{ASNQ} A large-scale AS2 corpus~\cite{garg2020tanda} derived from Google Natural Questions (NQ) dataset~\cite{kwiatkowski-etal-2019-natural}. It consists of ${\sim}$60K questions with labeled answer sentences. We use this as {\bf AS2} training data. 

\mypara{MS-MARCO NLG} A split of MS-MARCO~\cite{bajaj2018ms} that contains manually generated answers along with retrieved passages for ${\sim}$150k user queries, which we use for {\bf Fine-tuning}. We sub-sample 1k questions from the development set, along with their answer candidates extracted from the associated passages, to be used as \textbf{Evaluation} data in our experiments. (We do not use the entire development set of $\sim$ 100k questions for evaluation due to the expensive cost of human annotations).

\mypara{TREC-QA} A popular QA benchmark~\cite{wang2007jeopardy} used to evaluate AS2 models. For our experiments, we use the filtering and splits proposed in ~\cite{trecqazeyu}, where all questions have at least one positive and one negative candidate, and the test split is larger. The resulting dataset contains 816, 204 and 340 unique questions respectively for the training, dev.~and test sets.

\mypara{WikiQA} A popular AS2 dataset~\cite{yang2015wikiqa} containing questions from Bing search logs and answer candidates from Wikipedia. We use a `clean' setting for training by retaining questions with at least one positive answer candidate in the train and validation splits. This results in training/dev./test sets of WikiQA having 2118/296/236 questions, respectively.

\mypara{AQAD-L} The labeled counterpart of the industrial dataset AQAD-U as described in Section~\ref{ssec:unlabeled_data} above, where answer candidates additionally have human annotations of correctness/incorrectness. We use AQAD-L, comprising of ${\sim}$ 5k questions, as \textbf{Evaluation} data for experiments in the industrial setting. Results on AQAD-L are presented relative to the baseline AS2 model due to the data being internal. 

For data statistics, please refer to Appendix~\ref{ref:apx_data_statistics}.

\vspace{-.5em}
\subsection{Modeling Details}
\vspace{-.2em}
We use T5~\cite{raffeletall2020} as the model for GenQA {\MG}. For the AS2 models, we use a RoBERTa-Large~\cite{liu2019roberta} or ELECTRA-Base~\cite{clark2020electra} trained using the TANDA approach~\cite{garg2020tanda}, depending on the experimental setting. For our experiments, we set the value of $k{=}5$, i.e, the number of answer candidates to be provided as input to the GenQA model. 

We train our models using $fp16$ precision, Adam \cite{Kingma2015AdamAM} as optimizer with a $lr=1e{-}4$ and a batch size of $256$. We trained each model for $25$ epochs on both the versions of MS-MARCO (QA and NLG), and for $50$ epochs on WikiQA and TrecQA. We select the best model by maximizing the average AS2 score on the development set of each dataset instead of minimizing the validation loss (see the details in Appendix \ref{apx:ckpt_selection}).

\vspace{-.5em}
\subsection{Evaluation and Metrics}
\vspace{-.2em}
We perform human evaluation of our generated answers: for each question/answer pair, we collect the annotations from five annotators (corresponding to the answer being correct/incorrect) using Amazon MTurk (see Appendix~\ref{apx:maneval} for details). We use accuracy as the primary metric for all our experiments and models. Given a set of questions, this is computed as the fraction of correct answers divided by the number of incorrect answers as judged by the annotators. Note that: (i) each QA pair receives an average score from five annotators, and (ii) for the AS2 model, the accuracy is the same as Precision@1, which is the precision of the top ranked answer candidate.

\vspace{-.5em}
\section{Experiments and Results}
\vspace{-.2em}
\label{sec:experiments}
We perform experiments in three data settings to evaluate different features of our method. On the MS-MARCO datasets, we show that weak supervision can augment strong models trained on in-domain data. On WikiQA and TREC, we show that weak supervision on large data improves over direct supervision on small data for this QA task. We also present an experiment on a very large industrial dataset to measure the contribution of each of our proposed techniques for using unlabeled training data at scale.

{\color{red}
\begin{table*}
\begin{center}
    \resizebox{0.8\linewidth}{!}{
    \begin{tabular}{@{}ccccccccc@{}}
\toprule
\multirow{2}{*}{\textbf{Approach}}      & \multirow{2}{*}{\textbf{Model}}                                       & & \multicolumn{2}{c}{\textbf{Unlabeled: MS-MARCO QA}} & & \multicolumn{2}{c}{\textbf{Labeled: MS-MARCO NLG}}   & \multirow{2}{*}{\textbf{Accuracy (\%)}} \\ \cmidrule(lr){4-5} \cmidrule(lr){7-8}
                       &                                                               &  & \textbf{Used} & \textbf{Training Strategy \textbf{(Ours)}}     & & \textbf{Used} & \textbf{Training Strategy}   &                                \\ \midrule
\textbf{AS2}                    & \begin{tabular}[c]{@{}c@{}}RoBERTa-Large\end{tabular} & &{\xmark}    & -                     & & {\xmark}     & -                   & 79.3                           \\
\midrule
\multirow{5}{*}{\textbf{GenQA}} & T5 - Large                                                  &  & {\cmark}    & {\WS}                & & {\xmark}     & -                   & 79.9                           \\
                       & T5 - Large                                                    & & {\cmark}    & {\WS} + {\LW}         &  & {\xmark}     & -                   & 81.5                           \\
                       & T5 - Large                                                    & & {\cmark}    & {\WS} + {\SCI}        &  & {\xmark}     & -                   & 82.0                           \\
                       & T5 - Large                                                    & & {\cmark}    & {\WS} + {\SCO}        &  & {\xmark}     & -                   & 82.5                           \\
                       & T5 - Large                                                    & & {\cmark}    & {\WS} + {\LW} + {\SC} &  & {\xmark}     & -                   & 83.7                           \\
                       & T5 - Large                                                    & & {\xmark}    & -                     &  & {\cmark}     & ~\cite{hsu2021answer} & 82.6                           \\
                       & T5 - Large                                                    & & {\cmark}    & {\WS} + {\LW} + {\SC} &  &  {\cmark}     & ~\cite{hsu2021answer} & \textbf{85.3}                           \\ \bottomrule
\end{tabular}
    }
    \vspace{-0.5em}
    \caption{\small Results on the test split of MS MARCO-NLG for different training paradigms of GenQA models. The weak supervision is provided by a RoBERTa-Large AS2 model trained on ASNQ. We compare with a fully supervised GenQA baseline~\cite{hsu2021answer} trained on the train split of MS MARCO-NLG.}
    \label{tab:msmarco_results}
        \vspace{-1em}
\end{center}
\end{table*}
}

\vspace{-.5em}
\subsection{Comparison with the State of the art}
\vspace{-.2em}
These experiments aim at (i) understanding the effectiveness of our weakly supervised methods ({\WS}, {\LW} and {\SC}), and (ii) comparing them with a GenQA state-of-the-art model, which is trained using fully supervised training data. We create weakly supervised data with a RoBERTa-Large AS2 model trained on ASNQ by applying it to the MS-MARCO QA dataset. It is important to note that, in this setting, during the training, both the student and the teacher models do not have any knowledge of the original labels of MS-MARCO QA as we  consider this dataset to be unlabeled. We compare our approach against a supervised GenQA model baseline~\cite{hsu2021answer}, which is trained on MS-MARCO NLG. The MS-MARCO NLG dataset is much smaller than MS-MARCO QA but has higher quality answers as the targets since they are manually written. 
We also investigate a two-stage training strategy, by first applying our knowledge transfer approach, followed by the supervised training to understand if the two approaches are complementary or essentially capture similar information. All models are evaluated by manual annotators on the same MS-MARCO NLG test set. \vspace{-1em}

\paragraph{Results:} We present the results on the MS-MARCO NLG test set in Table~\ref{tab:msmarco_results}. The baseline zero-shot accuracy of the AS2 model on this data is 79.3 and the baseline accuracy of the fully supervised GenQA model~\cite{hsu2021answer} is 82.6. Our weak supervision ({\WS}) transfer technique, which does not use any answers written by annotators as targets, shows improvements over the AS2 baseline (0.6\%). This shows that our approach can transfer information learned by an AS2 model from ASNQ (a large labeled dataset) into a GenQA model.

Ablating each of the approaches ({\LW},{\SCI},{\SCO}) individually in addition to {\WS}, we observe consistent improvements (+1.6, +2.1 and +2.6\% respectively) over the performance of {\WS}, indicating that the AS2 scores can help in the knowledge transfer. Additionally, combining all the approaches with {\WS} significantly improves the performance, and surprisingly can even outperform the supervised GenQA baseline (by 1.1\% = 83.7{-}82.6). This shows that the knowledge transferred by our approach from ASNQ exceeds what can be learned from MS-MARCO NLG.\footnote{MS-MARCO NLG is larger than ASNQ, but the latter is a much higher quality dataset in terms of diversity and complexity of questions and answer annotations}

Finally, when we combine our weak supervised training techniques with the supervised training in a two stage pipeline, we observe very significant performance gains, e.g., 2.7\% over the supervised approach. This shows that (i) the information in MS-MARCO NLG is complementary to the knowledge transferred from ASNQ, and (ii) our approach is effective in  transferring knowledge from a discriminative ranker to a downstream GenQA model.

Due to brevity of space, we present a qualitative ablation of the generated examples in Appendix~\ref{apx:quality_eval}.

\begin{table}[t]
\begin{subtable}{\linewidth}
\centering
\resizebox{\linewidth}{!}{
   \begin{tabular}{@{}cccccccc@{}}
\toprule
\multirow{2}{*}{\textbf{Approach}}                                             & & \multicolumn{2}{c}{\textbf{Unlabeled (\small{MS-MARCO QA})}} & & \multicolumn{2}{c}{\textbf{Labeled (\small{WikiQA Train}) }}   & \multirow{2}{*}{\textbf{Accuracy (\%)}} \\ \cmidrule(lr){3-4} \cmidrule(lr){6-7}
                       &                                                               & \textbf{Used} & \textbf{Training Strategy}     & & \textbf{Used} & \textbf{Training Strategy}   &                                \\ \midrule
\textbf{AS2}                    &  &{\xmark}    & -                     & & {\cmark}     & Fine-tuning            & 78.3                           \\
\midrule
\multirow{5}{*}{\textbf{GenQA}}                                                &  & {\cmark}    & {\WS} + {\LW} + {\SC}               & & {\xmark}     & -                   & 78.7                           \\
                       
                                                                       & & {\xmark}    & -                     & & {\cmark}     & ~\cite{hsu2021answer} & 72.9                           \\
                                                                         & & {\cmark}    & {\WS} + {\LW} + {\SC} &&  {\cmark}     & ~\cite{hsu2021answer} & \textbf{79.8}                           \\ \bottomrule
\end{tabular}}
   \caption{WikiQA}\label{tab:wikiqa}
\end{subtable}

\bigskip

\begin{subtable}{\linewidth}
\centering
\resizebox{1\linewidth}{!}{
   \begin{tabular}{@{}cccccccc@{}}
\toprule
\multirow{2}{*}{\textbf{Approach}}                                       & & \multicolumn{2}{c}{\textbf{Unlabeled (\small{MS-MARCO QA})}} & & \multicolumn{2}{c}{\textbf{Labeled (\small{TREC-QA})}}   & \multirow{2}{*}{\textbf{Accuracy (\%)}} \\ \cmidrule(lr){3-4} \cmidrule(lr){6-7}
                       &                                                              & \textbf{Used} & \textbf{Training Strategy}     & & \textbf{Used} & \textbf{Training Strategy}   &                                \\ \midrule
\textbf{AS2}                    &  &{\xmark}    & -                     & & {\cmark}     & Fine-tuning            & 85.9                           \\
\midrule
\multirow{5}{*}{\textbf{GenQA}} 
&  & {\cmark} & {\WS} + {\LW} + {\SC} & & {\xmark} & - & \textbf{90.5} \\ 
&  & {\xmark} & - & & {\cmark} & ~\cite{hsu2021answer} & 80.7 \\
&  & {\cmark} & {\WS} + {\LW} + {\SC} && {\cmark} & ~\cite{hsu2021answer} & 89.8  \\ %TO CHANGE? 
\bottomrule
\end{tabular}}
   \caption{TREC-QA}\label{tab:trecqa}
\end{subtable}
\caption{\small Results on the test split of WikiQA and TREC-QA. The weak supervision is provided by RoBERTa-Large AS2 models trained respectively on WikiQA and TREC-QA. We compare with a fully supervised GenQA baseline~\cite{hsu2021answer} trained respectively on the train split of WikiQA and TREC-QA using ground truth correct answers as the target for generation. We use T5-Large for all GenQA models.} \label{tab:wiki_trec}
\vspace{-0.5em}
\end{table}
\vspace{-.6em}
\subsection{Scarce Data Setting}
\vspace{-.3em}
In this experiment, we measure the quality of our weak supervision approaches, by evaluating their performance on two popular AS2 benchmark datasets: WikiQA and TREC-QA. We train the AS2 teacher model on this data and still use the unlabeled data from MS-MARCO QA for performing the knowledge transfer. This way, we test if our approach is applicable in real scenarios, where data can be scarce and no large labeled data dataset is available (such as ASNQ or MS-MARCO NLG). Additionally, we verify if our approach works for other domains and if fine-tuning GenQA on the target domain data can help knowledge transfer in that domain, even in case of data scarcity.

We compare our weakly-supervised approaches with an AS2 baseline and a GenQA model trained on the target datasets, using their ground-truth labels. We used the original test splits of the datasets. Note that for these experiments, (i) we use the best performing strategy for our transfer learning, i.e.,  {\WS} along with {\LW} and {\SC}, and (ii) the AS2 baseline is the same model that we use to transfer knowledge on the MS-MARCO QA. \vspace{-.5em}

\paragraph{Results:} From our results in Table \ref{tab:wiki_trec}, we make the following observations:\\
(i) AS2 accuracy evaluated with our human annotations is around 10\% lower than results from previous works, e.g., \cite{zhang-etal-2021-joint}. As we use the same model,\footnote{Starting from the  \url{https://huggingface.co/roberta-large} checkpoint} the difference is due to the fact that we use the `raw' test setting which  includes questions with no correct answer candidates.\\ 
(ii) Our transfer learning techniques have better performance than both the AS2 model and the supervised GenQA baselines. For WikiQA, our knowledge transfer approach, which has only seen unlabeled MS-MARCO data and no labeled training data from WikiQA, gets higher accuracy than both the AS2 baseline (+0.4\%) and a fully supervised GenQA baseline (+5.8\%), which uses the ground truth labels from the target datasets. \\ 
(iii) We observe the same trend for TREC-QA: our weakly supervised models improve over both AS2 (4.6\%) and supervised GenQA (9.8\%) baselines.\\
(iv) In contrast to our observations from Table~\ref{tab:msmarco_results}, the supervised GenQA baseline for WikiQA and TREC-QA is less accurate than the AS2 baseline. We explain this with two reasons: (a) the small size of these datasets (only few thousands training questions) might be insufficient to train a large T5 model for GenQA, and (b) the usage of extracted answers as the target for generation instead of a human-written and natural sounding answer affects the quality of answer generation.\\
(v) Finally, supervised fine-tuning applied after our transfer learning only improves performance on WikiQA. The WikiQA dataset has several questions with no correct answers (${\sim}$40\%). Fine-tuning on the supervised dataset reinforces the training of the generator on questions having actual positive labels, thereby helping to reduce noise, and improving the final accuracy on the entire test set.

\vspace{-.5em}\subsection{Industrial Setting}

In this experiment, we aim to show that our experimental findings extend to very large-scale and real-world data, i.e., \emph{non-representative de-identified} customer questions from Alexa virtual assistant. We use the 50M question AQAD-U corpus as the unlabeled QA corpus for training the GenQA model, transferring the knowledge of an AS2 teacher model (no human-authored answer is used for training fully-supervised GenQA models on this data). We compare our methods for weak supervision against the AS2 teacher model on the labeled test split: AQAD-L.
\vspace{-.5em}

\paragraph{Results:}
We present the results in Table~\ref{tab:aqad_results} relative to the AS2 baseline, due to the data being internal, which is used as the `teacher' for transferring knowledge to train the GenQA model. For these experiments we use T5-Base as the GenQA model (due to the large size of AQAD-U), and our results show that knowledge transfer from the AS2 model using the unlabeled data surprisingly improves the accuracy of the baseline by 1.34\%. This indicates that the weak supervision provided by the AS2 model is able to train a GenQA model that performs better than the AS2 teacher itself. Furthermore, using loss weighting ({\LW}) and input/output shaping ({\SC}) significantly improves our weak supervision approach. The T5-Base model trained using a combination of {\LW} and {\SC} on the unlabeled AQAD-U corpus achieves an impressive 7.35\% gain in accuracy over the baseline AS2 model (which has been trained on labeled data with annotations for answer correctness). 

{\color{red}
\begin{table}
\begin{center}
    \resizebox{1\linewidth}{!}{
    % \begin{tabular}{lcccc} 
    %     \toprule
    %     Model & Teacher & Transfer & Supervised & Accuracy \\
    %     \toprule
    %     ELECTRA      &         &           &          &  AS2 baseline\\
    %     \midrule
    %     T5-base      & ELECTRA &  WS       &          &  +1.34\% \\
    %     T5-base      & ELECTRA & WS+LW     & LW       &  +4.08\%  \\
    %     T5-base      & ELECTRA & WS+LW+SCO & LW+SCO   &  \textbf{+7.35\%} \\
    %     %\midrule
    %     %T5-large     & ELECTRA & WS        & WS        &  +0.0\%  \\
    %     %T5-large     & ELECTRA & WS        & WS + LW   &  +3.2\%  \\
    %     %T5-large     & ELECTRA & WS        & WS+LW+SCO &  +4.8\%  \\
    %     %T5-large     & ELECTRA & WS+SCO    & WS+LW     &  +11.3\  \\
    %     %T5-large     & ELECTRA & WS+SCO    & WS+LW     &  +10.1\% \\
    % \end{tabular}
\begin{tabular}{@{}ccccccc@{}}
\toprule
\multirow{2}{*}{\textbf{Approach}}      & \multirow{2}{*}{\textbf{Model}}                                       & & \multicolumn{2}{c}{\textbf{Unlabeled: AQAD-U}} & &  \multirow{2}{*}{\textbf{Accuracy (\%)}} \\ \cmidrule(lr){4-5}
                       &                                                               &  & \textbf{Used} & \textbf{Training Strategy}     &                                \\ \midrule
\textbf{AS2}                    & \begin{tabular}[c]{@{}c@{}}ELECTRA-Base\end{tabular} & &{\xmark}    & -                     & &        Baseline        \\
\midrule
\multirow{3}{*}{\textbf{GenQA}} & T5 - Base                                                  &  & {\cmark}    & {\WS}                & &      +1.34\%                       \\
                       & T5 - Base                                                    & & {\cmark}    & {\WS} + {\LW}       &  &    +4.08\%                       \\
                       & T5 - Base                                                    & & {\cmark}    & {\WS} + {\LW} + {\SC} &          & \textbf{+7.35\%}                          \\ \bottomrule
\end{tabular}
 }
 \vspace{-0.75em}
    \caption{\small Results on $\text{AQAD}$-${\text{L}}$ for different training paradigms of GenQA models. All results are reported in absolute \% changes w.r.t the AS2 baseline.}
    \label{tab:aqad_results}
    \vspace{-1em}
\end{center}
\end{table}
}

\vspace{-.5em}
\section{Analysis and Ablation Studies}
\vspace{-.2em}
\label{sec:analysis}

\paragraph{Automatic Evaluation:} We consider whether automatic evaluation metrics correlate with human evaluation for our task. In Table \ref{tab:automaticmetrics}, we compare BERT-Score~\cite{Zhang*2020BERTScore:} and BLEURT~\cite{sellam-etal-2020-bleurt} with the human  evaluation of various models on MS MARCO-NLG test data. We find that despite their good performance for other NLG tasks neither metric has a particularly strong Pearson correlation with the human evaluation results for the task of answer sentence generation (GenQA): BLEURT has a correlation 0.622; BERT-Score has a correlation of 0.447. Neither automatic metric is able to correctly identify the system ranking effected by human evaluation as presented in Table~\ref{tab:msmarco_results}. Additional analysis using the BLEU score is presented in Appendix~\ref{apx:acc_vs_bleu}.

\begin{table}
\tiny
\begin{center}
\resizebox{\linewidth}{!}{
\begin{tabular}{@{}ccccccccc@{}}
\toprule
\multirow{2}{*}{\textbf{Approach/Strategy}} & \textbf{Human } && \multicolumn{3}{c}{\textbf{BERT-Score}}   && \multicolumn{2}{c}{\textbf{BLEURT}}  \\
  & \textbf{Evaluation} && \textbf{PREC} & \textbf{REC} & \textbf{F1} & & \textbf{AVG} & \textbf{STDEV} \\
\midrule
 \textbf{AS2}                                    & 0.793 && 0.876 & 0.905 & 0.890 && 0.509 & 0.225     \\
 \textbf{WS}                            & 0.799 && 0.884 & 0.903 & 0.893 && 0.520 & 0.217    \\
                                    \textbf{WS+LW}                            & 0.815 && 0.993 & 0.994 & 0.993 && 0.517 & 0.216     \\
                                    \textbf{WS+SCI}                           & 0.820 && 0.885 & 0.905 & 0.895 && 0.525 & 0.218     \\
                                \textbf{WS+SCO}                           & 0.825 && 0.884 & 0.905 & 0.894 && 0.521 & 0.221     \\
                                   \textbf{WS+LW+SC}                         & 0.837 && 0.883 & 0.901 & 0.891 && 0.512 & 0.211    \\
                                    \cite{hsu2021answer}                   & 0.826 && 0.907 & 0.911 & 0.908 && 0.555 & 0.224     \\
                                 \textbf{WS+LW+SC}+   & \multirow{2}{*}{0.853} && \multirow{2}{*}{0.994} & \multirow{2}{*}{0.996} & \multirow{2}{*}{0.995} && \multirow{2}{*}{0.559} & \multirow{2}{*}{0.221}     \\
                                  \cite{hsu2021answer} & &&&&&&\\
\midrule
\multicolumn{2}{c}{\textbf{Correlation (with Human Evaluation)}} && \multicolumn{3}{c}{\textbf{0.447}} && \multicolumn{2}{c}{\textbf{0.622}} \\
\bottomrule
\end{tabular}}
\vspace{-0.75em}
\caption{\small Results on the the testset of MS MARCO-NLG comparing human evaluation to well-known automatic evaluation metrics: BERT-Score and BLEURT. The last row shows the correlation between the human evaluation and the two automatic metrics (BERT-Score, and BLEURT), indicating that these metrics do not correlate strongly with human evaluation.}
\label{tab:automaticmetrics}
% \vspace{-1em}
\end{center}
\end{table}

%Model	Human Evaluation	BERT-SCORE	BLEURT	Correlation (BERT-Score, Human)	Correlation (BLEURT, Human)
%Pr	Re	F1	Avg	Std. Dev.
% roberta-large	& 0.793 & 0.87605 & 0.90452 & 0.88957 & 0.50947 & 0.22461 & 0.44745 & 0.62195
%t5-large-ws 	&	0.799		& 0.88422 	&	0.90323 	&	0.89315 	&	0.51951 	&	0.21706
%t5-large-lw 	&	0.815 	&	0.99253 	&	0.99408 	&	0.9933 	&	0.51732 	&	0.21617
%t5-large-sci 	&	0.82 	&	0.88527 	&	0.90535 	&	0.8947 	&	0.52469 	&	0.21765
%t5-large-sco 	&	0.825	0.88361 	&	0.90461 	&	0.89351 	&	0.52147 	&	0.22129
%t5-large-lw+sci+sco 	&	0.837 	&	0.8827 	&	0.90141 	&	0.89146 	&	0.51166 	&	0.2107
%t5-large-genqa 	&	0.826 	&	0.90696 	&	0.91071 	&	0.90848 	&	0.55517 	&	0.22446
%t5-large-lw+sci+sco+ft-genqa 	&	0.853 	&	0.99359 	&	0.99575 	&	0.99467 	&	0.55857 	&	0.22084

\paragraph{Structured Output and Accuracy:} In the {\SCO} approach we propose, we prepend a special bucket token $b_1 = \mathcal{F}(q,c_{M_1})$ corresponding to the AS2 confidence score of the target answer candidate $c_{M_1}$, so that the output target for the GenQA model {\MG} is: ``$b_1 \ c_{M_1}$''. We denote the bucket tokens for $l{=}5$ with the following set: [{\tiny{\YES},{\PROBABLY},{\MAYBE}, {\DOUBT},{\NO}}] corresponding to the confidence intervals $[0.8,1]$, $[0.6,0.8)$, $[0.4, 0,6)$, $[0.2,0.4)$ and $[0,0.2)$ for {\MA}'s score respectively. In this section, we analyze the role of the {\SCO} bucket tokens as a confidence measure for the GenQA model's output during inference. For this, we cluster the answers generated by the GenQA model (a T5-Large model trained on MS-MARCO QA using both {\LW} and {\SC}) based on the generated {\SCO} bucket token. We then manually evaluate the answers and show the accuracy for each cluster in Table~\ref{tab:sco_confidence_greedy}. Here we observe an evident difference in the correctness of the generated answers in the {\YES} cluster from those in the {\NO} cluster (78.6\% v/s 60.5\%). The accuracy of generated answers monotonically decreases with the confidence interval of the bucket token ({\YES} $\rightarrow$ {\NO}). Additional analysis of model outputs is presented in Appendix~\ref{apx:quality_eval}.

%\begin{table}[h]
%\small
%\begin{center}
%    \begin{tabular}{lc} 
%        \toprule
%        Starting Token & Accuracy \\
%        \toprule
%        $[\_\text{YES}\_]$      & 0.799 \\
%        $[\_\text{PROBABLY}\_]$ & 0.723 \\
%        $[\_\text{MAYBE}\_]$    & 0.571 \\
%        $[\_\text{DOUBT}\_]$    & 0.698 \\
%        $[\_\text{NO}\_]$       & 0.55 \\
%       \bottomrule
%    \end{tabular}
%    \caption{with beam search}
%    \label{tab:sco_confidence_beam}
%\end{center}
%\end{table}

\begin{table}[t]
\small
\begin{center}
\resizebox{0.65\linewidth}{!}{
    \begin{tabular}{lccc} 
        \toprule
        Starting {\SCO} Token & Accuracy \% \\% & BLEU(gen vs rr) \\
        \toprule
        \tiny{\YES}      & 78.6 \\%& 24.84  \\
        \tiny{\PROBABLY} & 71.2 \\%& 25.51  \\
        \tiny{\MAYBE}    & 69.1 \\%& 25.93  \\
        \tiny{\DOUBT}    & 66.0 \\%& 25.24  \\
        \tiny{\NO}       & 60.5 \\%& 21.71  \\
       \bottomrule
    \end{tabular}
}
\vspace{-0.75em}
    \caption{\small Accuracy on generated answers clustered according to the starting {\SCO} bucket token generated by the GenQA model on the WikiQA test set}
    \label{tab:sco_confidence_greedy}
\end{center}
\end{table}

%\begin{table}[h]
%\small
%\begin{center}
%    \begin{tabular}{lccccc} 
%        \toprule
%        Starting Token & C1 & C2 & C3 & C4 & All \\
%        \toprule
%            YES      & 24.84 & 18.62 & 17.76 & 17.26 & 50.62 \\
%            PROBABLY & 25.50 & 19.32 & 18.07 & 18.95 & 52.70 \\
%            MAYBE    & 25.93 & 20.84 & 19.84 & 18.13 & 53.76 \\
%            DOUBT    & 25.24 & 22.03 & 21.84 & 17.47 & 48.93 \\
%            NO       & 21.71 & 21.90 & 22.40 & 22.74 & 51.50 \\
%       \bottomrule
%    \end{tabular}
%    \caption{BLEU Score from the generated answer and the separate input candidates}
%    \label{tab:bleu_vs_inputs}
%\end{center}
%\end{table}
\begin{figure}
  \centering
  \begin{tikzpicture}[scale=0.8,font=\small]
    \begin{axis}[
      width=0.95\linewidth,
      ybar,
      bar width=5pt,
      xlabel={Input candidates},
      ylabel={BLEU},
      ymin=0,
      xtick=data,
      axis x line=bottom,
      axis y line=left,
      enlarge x limits=0.15,
      symbolic x coords={c1,c2,c3,c4,all},
      legend columns=-1,
      legend style={at={(0.5,1.16)},anchor=north,legend cell align=left, nodes={scale=0.5, transform shape,draw=none}}
    ]
      \addplot[fill=green] coordinates {
            (c1,24.84)
            (c2,18.62)
            (c3,17.76)
            (c4,17.26)
            %(all,50.62)
        };
        \addplot[fill=yellow] coordinates {
            (c1,25.50)
            (c2,19.32)
            (c3,18.07)
            (c4,18.95)
            %(all,52.70)
        };
        \addplot[fill=orange] coordinates {
            (c1,25.93)
            (c2,20.84)
            (c3,19.84)
            (c4,18.13)
            %(all,53.76)
        };
        \addplot[fill=magenta] coordinates {
            (c1,25.24)
            (c2,22.03)
            (c3,21.84)
            (c4,17.47)
            %(all,48.93)
        };
        \addplot[fill=red] coordinates {
            (c1,21.71)
            (c2,21.90)
            (c3,22.40)
            (c4,22.74)
            %(all,51.50)
        };
    \legend{\YES,\PROBABLY,\MAYBE,\DOUBT,\NO}
    \end{axis}
    \end{tikzpicture}
    \vspace{-1em}
    \caption{\small Correlation between the generated answer (with different starting {\SCO} bucket tokens) with each input candidate. We observe that generated answers starting with high confidence bucket tokens (e.g., {{\YES}} and {{\PROBABLY}}) tend to copy more from the top ranked answer candidate, than the generated answers starting with lower confidence bucket tokens (e.g., {{\NO}}).}
    \label{fig:bleu_generated_vs_candidates}
\end{figure}
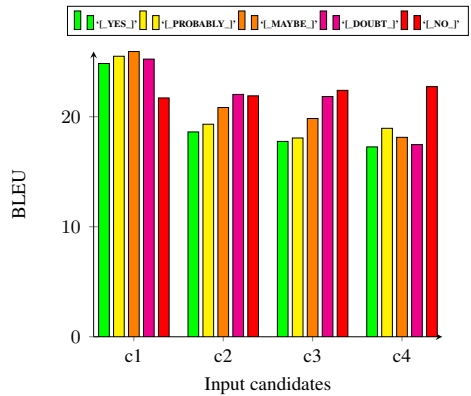

\paragraph{Generated Answer v/s Input Candidates:} We compare the generated answers (for each {\SCO} bucket token) with the input answer candidates to understand how the model copies from the input candidates, and if there is a correlation with the {\SCO} bucket tokens. In Fig.~\ref{fig:bleu_generated_vs_candidates}, we present the similarity between the generated answer with the top $4$ ranked input candidates using BLEU score. This analysis shows that generated answers starting with a high confidence {\SCO} bucket token (e.g., {{\YES}} or {{\PROBABLY}}) are more similar to the first candidate (higher ranked), while answers starting with lower confidence {\SCO} bucket tokens (e.g., {{\NO}}) are on average equally distant from all the input candidates.

\vspace{-.5em}
\section{Conclusion}
\vspace{-.2em}
\label{sec:conclusion}
In this paper, we have presented a novel approach for transferring knowledge from a discriminative AS2 model to an answer generation model, only using unlabeled data. We use the ranking produced by the AS2 model for training a GenQA model using the top answer as the target output, and the next $k$ top ranked answers along with the question as the input. We also propose input/output shaping and loss weighting techniques during knowledge transfer to improve the performance of GenQA. Our experimental results on three public and one large industrial datasets show that GenQA models trained with knowledge transfer from  AS2 models achieve higher answering accuracy than both the AS2 teacher and supervised GenQA trained with in-domain data. We are releasing our code and trained models to support future research.

\section*{Limitations}
Our approach of training GenQA models requires access to large GPU resources for training large pre-trained language models such as T5-Large, etc. For the experiments in this paper, we only consider datasets from the English language, however we conjecture that our techniques should work similarly for languages with a similar morphology. The evaluations for all experiments performed in this paper are done using human annotations on MTurk, which is time consuming and expensive. Currently, automatic evaluation of correctness and style suitability for question answering is extremely challenging, and we hope that research advances in this domain further encourages broader research in answer generation systems.

\section*{Acknowledgements}
We thank the anonymous reviewers and the meta-reviewer for their valuable suggestions and comments. We would like to thank Thuy Vu for developing and sharing the AQAD dataset.

\bibliography{anthology,custom}

\begin{thebibliography}{49}
\expandafter\ifx\csname natexlab\endcsname\relax\def\natexlab#1{#1}\fi

\bibitem[{Asai et~al.(2022)Asai, Gardner, and
  Hajishirzi}]{DBLP:journals/corr/abs-2112-08688}
Akari Asai, Matt Gardner, and Hannaneh Hajishirzi. 2022.
\newblock \href {https://doi.org/10.18653/v1/2022.naacl-main.162}
  {Evidentiality-guided generation for knowledge-intensive {NLP} tasks}.
\newblock In \emph{Proceedings of the 2022 Conference of the North American
  Chapter of the Association for Computational Linguistics: Human Language
  Technologies}, pages 2226--2243, Seattle, United States. Association for
  Computational Linguistics.

\bibitem[{Bajaj et~al.(2018)Bajaj, Campos, Craswell, Deng, Gao, Liu, Majumder,
  McNamara, Mitra, Nguyen, Rosenberg, Song, Stoica, Tiwary, and
  Wang}]{bajaj2018ms}
Payal Bajaj, Daniel Campos, Nick Craswell, Li~Deng, Jianfeng Gao, Xiaodong Liu,
  Rangan Majumder, Andrew McNamara, Bhaskar Mitra, Tri Nguyen, Mir Rosenberg,
  Xia Song, Alina Stoica, Saurabh Tiwary, and Tong Wang. 2018.
\newblock \href {http://arxiv.org/abs/1611.09268} {Ms marco: A human generated
  machine reading comprehension dataset}.

\bibitem[{Bian et~al.(2017)Bian, Li, Yang, Chen, and Lin}]{Bian2017ACM}
Weijie Bian, Si~Li, Zhao Yang, Guang Chen, and Zhiqing Lin. 2017.
\newblock A compare-aggregate model with dynamic-clip attention for answer
  selection.
\newblock \emph{Proceedings of the 2017 ACM on Conference on Information and
  Knowledge Management}.

\bibitem[{Chen et~al.(2017)Chen, Fisch, Weston, and Bordes}]{Chen-Fisch-2017}
Danqi Chen, Adam Fisch, Jason Weston, and Antoine Bordes. 2017.
\newblock \href {https://doi.org/10.18653/v1/P17-1171} {Reading {W}ikipedia to
  answer open-domain questions}.
\newblock In \emph{Proceedings of the 55th Annual Meeting of the Association
  for Computational Linguistics (Volume 1: Long Papers)}, pages 1870--1879,
  Vancouver, Canada. Association for Computational Linguistics.

\bibitem[{Clark et~al.(2020)Clark, Luong, Le, and Manning}]{clark2020electra}
Kevin Clark, Minh-Thang Luong, Quoc~V. Le, and Christopher~D. Manning. 2020.
\newblock \href {https://openreview.net/pdf?id=r1xMH1BtvB} {{ELECTRA}:
  Pre-training text encoders as discriminators rather than generators}.
\newblock In \emph{ICLR}.

\bibitem[{Deng et~al.(2020)Deng, Lam, Xie, Chen, Li, Yang, and
  Shen}]{Deng2020JointLO}
Yang Deng, Wai Lam, Yuexiang Xie, Daoyuan Chen, Yaliang Li, Min Yang, and Ying
  Shen. 2020.
\newblock Joint learning of answer selection and answer summary generation in
  community question answering.
\newblock In \emph{AAAI}.

\bibitem[{Devlin et~al.(2019)Devlin, Chang, Lee, and
  Toutanova}]{devlinetal2019}
Jacob Devlin, Ming-Wei Chang, Kenton Lee, and Kristina Toutanova. 2019.
\newblock \href {https://doi.org/10.18653/v1/N19-1423} {{BERT}: Pre-training of
  deep bidirectional transformers for language understanding}.
\newblock In \emph{Proceedings of the 2019 Conference of the North {A}merican
  Chapter of the Association for Computational Linguistics: Human Language
  Technologies, Volume 1 (Long and Short Papers)}, pages 4171--4186,
  Minneapolis, Minnesota. Association for Computational Linguistics.

\bibitem[{Di~Liello et~al.(2022{\natexlab{a}})Di~Liello, Garg, Soldaini, and
  Moschitti}]{di-liello-etal-2022-paragraph}
Luca Di~Liello, Siddhant Garg, Luca Soldaini, and Alessandro Moschitti.
  2022{\natexlab{a}}.
\newblock \href {https://doi.org/10.18653/v1/2022.naacl-main.181}
  {Paragraph-based transformer pre-training for multi-sentence inference}.
\newblock In \emph{Proceedings of the 2022 Conference of the North American
  Chapter of the Association for Computational Linguistics: Human Language
  Technologies}, pages 2521--2531, Seattle, United States. Association for
  Computational Linguistics.

\bibitem[{Di~Liello et~al.(2022{\natexlab{b}})Di~Liello, Garg, Soldaini, and
  Moschitti}]{diliello2021pretraininingtransformersas2}
Luca Di~Liello, Siddhant Garg, Luca Soldaini, and Alessandro Moschitti.
  2022{\natexlab{b}}.
\newblock \href {https://doi.org/10.48550/ARXIV.2205.10455} {Pre-training
  transformer models with sentence-level objectives for answer sentence
  selection}.

\bibitem[{Fajcik et~al.(2021)Fajcik, Docekal, Ondrej, and
  Smrz}]{fajcik-etal-2021-r2-d2}
Martin Fajcik, Martin Docekal, Karel Ondrej, and Pavel Smrz. 2021.
\newblock \href {https://doi.org/10.18653/v1/2021.findings-emnlp.73} {{R2-D2}:
  A modular baseline for open-domain question answering}.
\newblock In \emph{Findings of the Association for Computational Linguistics:
  EMNLP 2021}, pages 854--870, Punta Cana, Dominican Republic. Association for
  Computational Linguistics.

\bibitem[{Garg and Moschitti(2021)}]{garg-moschitti-2021-will}
Siddhant Garg and Alessandro Moschitti. 2021.
\newblock \href {https://doi.org/10.18653/v1/2021.emnlp-main.583} {Will this
  question be answered? question filtering via answer model distillation for
  efficient question answering}.
\newblock In \emph{Proceedings of the 2021 Conference on Empirical Methods in
  Natural Language Processing}, pages 7329--7346, Online and Punta Cana,
  Dominican Republic. Association for Computational Linguistics.

\bibitem[{Garg et~al.(2020)Garg, Vu, and Moschitti}]{garg2020tanda}
Siddhant Garg, Thuy Vu, and Alessandro Moschitti. 2020.
\newblock \href {https://doi.org/10.1609/aaai.v34i05.6282} {Tanda: Transfer and
  adapt pre-trained transformer models for answer sentence selection}.
\newblock \emph{Proceedings of the AAAI Conference on Artificial Intelligence},
  34(05):7780--7788.

\bibitem[{Goodwin et~al.(2020)Goodwin, Savery, and
  Demner-Fushman}]{Goodwin2020TowardsZS}
Travis Goodwin, Max~E. Savery, and Dina Demner-Fushman. 2020.
\newblock Towards zero shot conditional summarization with adaptive multi-task
  fine-tuning.
\newblock \emph{Proceedings of the Conference on Empirical Methods in Natural
  Language Processing. Conference on Empirical Methods in Natural Language
  Processing}, 2020:3215--3226.

\bibitem[{Gururangan et~al.(2020)Gururangan, Marasovi{\'c}, Swayamdipta, Lo,
  Beltagy, Downey, and Smith}]{Gururangan2020DontSP}
Suchin Gururangan, Ana Marasovi{\'c}, Swabha Swayamdipta, Kyle Lo, Iz~Beltagy,
  Doug Downey, and Noah~A. Smith. 2020.
\newblock \href {https://doi.org/10.18653/v1/2020.acl-main.740} {Don{'}t stop
  pretraining: Adapt language models to domains and tasks}.
\newblock In \emph{Proceedings of the 58th Annual Meeting of the Association
  for Computational Linguistics}, pages 8342--8360, Online. Association for
  Computational Linguistics.

\bibitem[{Hinton et~al.(2015)Hinton, Vinyals, and Dean}]{hinton2015distilling}
Geoffrey Hinton, Oriol Vinyals, and Jeff Dean. 2015.
\newblock \href {http://arxiv.org/abs/1503.02531} {Distilling the knowledge in
  a neural network}.

\bibitem[{Hsu et~al.(2021)Hsu, Lind, Soldaini, and Moschitti}]{hsu2021answer}
Chao-Chun Hsu, Eric Lind, Luca Soldaini, and Alessandro Moschitti. 2021.
\newblock \href {https://doi.org/10.18653/v1/2021.findings-acl.374} {Answer
  generation for retrieval-based question answering systems}.
\newblock In \emph{Findings of the Association for Computational Linguistics:
  ACL-IJCNLP 2021}, pages 4276--4282, Online. Association for Computational
  Linguistics.

\bibitem[{Iida et~al.(2019)Iida, Kruengkrai, Ishida, Torisawa, Oh, and
  Kloetzer}]{Iida2019ExploitingBK}
Ryu Iida, Canasai Kruengkrai, Ryo Ishida, Kentaro Torisawa, Jong-Hoon Oh, and
  Julien Kloetzer. 2019.
\newblock Exploiting background knowledge in compact answer generation for
  why-questions.
\newblock In \emph{AAAI}.

\bibitem[{Izacard and Grave(2021)}]{Izacard2021LeveragingPR}
Gautier Izacard and Edouard Grave. 2021.
\newblock \href {https://doi.org/10.18653/v1/2021.eacl-main.74} {Leveraging
  passage retrieval with generative models for open domain question answering}.
\newblock In \emph{Proceedings of the 16th Conference of the European Chapter
  of the Association for Computational Linguistics: Main Volume}, pages
  874--880, Online. Association for Computational Linguistics.

\bibitem[{Kamath et~al.(2020)Kamath, Jia, and
  Liang}]{kamath-etal-2020-selective}
Amita Kamath, Robin Jia, and Percy Liang. 2020.
\newblock \href {https://doi.org/10.18653/v1/2020.acl-main.503} {Selective
  question answering under domain shift}.
\newblock In \emph{Proceedings of the 58th Annual Meeting of the Association
  for Computational Linguistics}, pages 5684--5696, Online. Association for
  Computational Linguistics.

\bibitem[{Kingma and Ba(2015)}]{Kingma2015AdamAM}
Diederik~P. Kingma and Jimmy Ba. 2015.
\newblock \href {http://dblp.uni-trier.de/db/conf/iclr/iclr2015.html#KingmaB14}
  {Adam: A method for stochastic optimization.}
\newblock In \emph{International Conference of Learning Representations}.

\bibitem[{Kwiatkowski et~al.(2019)Kwiatkowski, Palomaki, Redfield, Collins,
  Parikh, Alberti, Epstein, Polosukhin, Devlin, Lee, Toutanova, Jones, Kelcey,
  Chang, Dai, Uszkoreit, Le, and Petrov}]{kwiatkowski-etal-2019-natural}
Tom Kwiatkowski, Jennimaria Palomaki, Olivia Redfield, Michael Collins, Ankur
  Parikh, Chris Alberti, Danielle Epstein, Illia Polosukhin, Jacob Devlin,
  Kenton Lee, Kristina Toutanova, Llion Jones, Matthew Kelcey, Ming-Wei Chang,
  Andrew~M. Dai, Jakob Uszkoreit, Quoc Le, and Slav Petrov. 2019.
\newblock \href {https://doi.org/10.1162/tacl_a_00276} {Natural questions: A
  benchmark for question answering research}.
\newblock \emph{Transactions of the Association for Computational Linguistics},
  7:452--466.

\bibitem[{Lewis et~al.(2020{\natexlab{a}})Lewis, Liu, Goyal, Ghazvininejad,
  Mohamed, Levy, Stoyanov, and Zettlemoyer}]{lewis-etal-2020-bart}
Mike Lewis, Yinhan Liu, Naman Goyal, Marjan Ghazvininejad, Abdelrahman Mohamed,
  Omer Levy, Veselin Stoyanov, and Luke Zettlemoyer. 2020{\natexlab{a}}.
\newblock \href {https://doi.org/10.18653/v1/2020.acl-main.703} {{BART}:
  Denoising sequence-to-sequence pre-training for natural language generation,
  translation, and comprehension}.
\newblock In \emph{Proceedings of the 58th Annual Meeting of the Association
  for Computational Linguistics}, pages 7871--7880, Online. Association for
  Computational Linguistics.

\bibitem[{Lewis et~al.(2020{\natexlab{b}})Lewis, Perez, Piktus, Petroni,
  Karpukhin, Goyal, K\"{u}ttler, Lewis, Yih, Rockt\"{a}schel, Riedel, and
  Kiela}]{10.5555/3495724.3496517}
Patrick Lewis, Ethan Perez, Aleksandra Piktus, Fabio Petroni, Vladimir
  Karpukhin, Naman Goyal, Heinrich K\"{u}ttler, Mike Lewis, Wen-tau Yih, Tim
  Rockt\"{a}schel, Sebastian Riedel, and Douwe Kiela. 2020{\natexlab{b}}.
\newblock Retrieval-augmented generation for knowledge-intensive nlp tasks.
\newblock In \emph{Proceedings of the 34th International Conference on Neural
  Information Processing Systems}, NIPS'20, Red Hook, NY, USA. Curran
  Associates Inc.

\bibitem[{Liu et~al.(2019)Liu, Ott, Goyal, Du, Joshi, Chen, Levy, Lewis,
  Zettlemoyer, and Stoyanov}]{liu2019roberta}
Yinhan Liu, Myle Ott, Naman Goyal, Jingfei Du, Mandar Joshi, Danqi Chen, Omer
  Levy, Mike Lewis, Luke Zettlemoyer, and Veselin Stoyanov. 2019.
\newblock \href {http://arxiv.org/abs/1907.11692} {Roberta: A robustly
  optimized bert pretraining approach}.

\bibitem[{Luong et~al.(2016)Luong, Le, Sutskever, Vinyals, and
  Kaiser}]{luong2015multi}
Minh{-}Thang Luong, Quoc~V. Le, Ilya Sutskever, Oriol Vinyals, and Lukasz
  Kaiser. 2016.
\newblock \href {http://arxiv.org/abs/1511.06114} {Multi-task sequence to
  sequence learning}.
\newblock In \emph{4th International Conference on Learning Representations,
  {ICLR} 2016, San Juan, Puerto Rico, May 2-4, 2016, Conference Track
  Proceedings}.

\bibitem[{Matsubara et~al.(2020{\natexlab{a}})Matsubara, Vu, and
  Moschitti}]{DBLP:conf/sigir/MatsubaraVM20}
Yoshitomo Matsubara, Thuy Vu, and Alessandro Moschitti. 2020{\natexlab{a}}.
\newblock \href {https://doi.org/10.1145/3397271.3401266} {Reranking for
  efficient transformer-based answer selection}.
\newblock In \emph{Proceedings of the 43rd International {ACM} {SIGIR}
  conference on research and development in Information Retrieval, {SIGIR}
  2020, Virtual Event, China, July 25-30, 2020}, pages 1577--1580. {ACM}.

\bibitem[{Matsubara et~al.(2020{\natexlab{b}})Matsubara, Vu, and
  Moschitti}]{Matsubara2020RerankingFE}
Yoshitomo Matsubara, Thuy Vu, and Alessandro Moschitti. 2020{\natexlab{b}}.
\newblock \href {https://doi.org/10.1145/3397271.3401266} {Reranking for
  efficient transformer-based answer selection}.
\newblock In \emph{Proceedings of the 43rd International ACM SIGIR Conference
  on Research and Development in Information Retrieval}, SIGIR '20, page
  1577–1580, New York, NY, USA. Association for Computing Machinery.

\bibitem[{Muller et~al.(2021)Muller, Soldaini, Koncel{-}Kedziorski, Lind, and
  Moschitti}]{DBLP:journals/corr/abs-2110-07150}
Benjamin Muller, Luca Soldaini, Rik Koncel{-}Kedziorski, Eric Lind, and
  Alessandro Moschitti. 2021.
\newblock \href {http://arxiv.org/abs/2110.07150} {Cross-lingual genqa: {A}
  language-agnostic generative question answering approach for open-domain
  question answering}.
\newblock \emph{CoRR}, abs/2110.07150.

\bibitem[{Radford et~al.(2019)Radford, Wu, Child, Luan, Amodei, and
  Sutskever}]{radfordetall2019gpt2}
Alec Radford, Jeff Wu, Rewon Child, David Luan, Dario Amodei, and Ilya
  Sutskever. 2019.
\newblock Language models are unsupervised multitask learners.

\bibitem[{Raffel et~al.(2020)Raffel, Shazeer, Roberts, Lee, Narang, Matena,
  Zhou, Li, and Liu}]{raffeletall2020}
Colin Raffel, Noam Shazeer, Adam Roberts, Katherine Lee, Sharan Narang, Michael
  Matena, Yanqi Zhou, Wei Li, and Peter~J. Liu. 2020.
\newblock \href {http://jmlr.org/papers/v21/20-074.html} {Exploring the limits
  of transfer learning with a unified text-to-text transformer}.
\newblock \emph{Journal of Machine Learning Research}, 21(140):1--67.

\bibitem[{Sanh et~al.(2019)Sanh, Debut, Chaumond, and
  Wolf}]{sanh2020distilbert}
Victor Sanh, Lysandre Debut, Julien Chaumond, and Thomas Wolf. 2019.
\newblock \href {http://arxiv.org/abs/1910.01108} {{DistilBERT, a distilled
  version of BERT: smaller, faster, cheaper and lighter}}.
\newblock In \emph{5th Workshop on Energy Efficient Machine Learning and
  Cognitive Computing @ NeurIPS 2019}.

\bibitem[{Schuster et~al.(2019)Schuster, Gupta, Shah, and
  Lewis}]{Schuster2019CrosslingualTL}
Sebastian Schuster, Sonal Gupta, Rushin Shah, and Mike Lewis. 2019.
\newblock \href {https://doi.org/10.18653/v1/N19-1380} {Cross-lingual transfer
  learning for multilingual task oriented dialog}.
\newblock In \emph{Proceedings of the 2019 Conference of the North {A}merican
  Chapter of the Association for Computational Linguistics: Human Language
  Technologies, Volume 1 (Long and Short Papers)}, pages 3795--3805,
  Minneapolis, Minnesota. Association for Computational Linguistics.

\bibitem[{Sellam et~al.(2020)Sellam, Das, and Parikh}]{sellam-etal-2020-bleurt}
Thibault Sellam, Dipanjan Das, and Ankur Parikh. 2020.
\newblock \href {https://doi.org/10.18653/v1/2020.acl-main.704} {{BLEURT}:
  Learning robust metrics for text generation}.
\newblock In \emph{Proceedings of the 58th Annual Meeting of the Association
  for Computational Linguistics}, pages 7881--7892, Online. Association for
  Computational Linguistics.

\bibitem[{Severyn and Moschitti(2015)}]{Severyn2015LearningTR}
Aliaksei Severyn and Alessandro Moschitti. 2015.
\newblock Learning to rank short text pairs with convolutional deep neural
  networks.
\newblock \emph{Proceedings of the 38th International ACM SIGIR Conference on
  Research and Development in Information Retrieval}.

\bibitem[{Shazeer and Stern(2018)}]{Shazeer2018AdafactorAL}
Noam~M. Shazeer and Mitchell Stern. 2018.
\newblock Adafactor: Adaptive learning rates with sublinear memory cost.
\newblock \emph{ArXiv}, abs/1804.04235.

\bibitem[{Shen et~al.(2017)Shen, Yang, and Deng}]{Shen2017InterWeightedAN}
Gehui Shen, Yunlun Yang, and Zhi-Hong Deng. 2017.
\newblock \href {https://doi.org/10.18653/v1/D17-1122} {Inter-weighted
  alignment network for sentence pair modeling}.
\newblock In \emph{Proceedings of the 2017 Conference on Empirical Methods in
  Natural Language Processing}, pages 1179--1189, Copenhagen, Denmark.
  Association for Computational Linguistics.

\bibitem[{Tay et~al.(2018)Tay, Luu, and Hui}]{Tay2018MultiCastAN}
Yi~Tay, Anh~Tuan Luu, and Siu~Cheung Hui. 2018.
\newblock Multi-cast attention networks.
\newblock \emph{Proceedings of the 24th ACM SIGKDD International Conference on
  Knowledge Discovery \& Data Mining}.

\bibitem[{Tayyar~Madabushi et~al.(2018)Tayyar~Madabushi, Lee, and
  Barnden}]{Madabushi2018IntegratingQC}
Harish Tayyar~Madabushi, Mark Lee, and John Barnden. 2018.
\newblock \href {https://aclanthology.org/C18-1278} {Integrating question
  classification and deep learning for improved answer selection}.
\newblock In \emph{Proceedings of the 27th International Conference on
  Computational Linguistics}, pages 3283--3294, Santa Fe, New Mexico, USA.
  Association for Computational Linguistics.

\bibitem[{Tran et~al.(2018)Tran, Lai, Haffari, Zukerman, Bui, and
  Bui}]{Tran2018TheCA}
Quan~Hung Tran, Tuan Lai, Gholamreza Haffari, Ingrid Zukerman, Trung Bui, and
  Hung Bui. 2018.
\newblock \href {https://doi.org/10.18653/v1/N18-1115} {The context-dependent
  additive recurrent neural net}.
\newblock In \emph{Proceedings of the 2018 Conference of the North {A}merican
  Chapter of the Association for Computational Linguistics: Human Language
  Technologies, Volume 1 (Long Papers)}, pages 1274--1283, New Orleans,
  Louisiana. Association for Computational Linguistics.

\bibitem[{Voorhees(1999)}]{voorhees99trec}
Ellen~M. Voorhees. 1999.
\newblock The trec-8 question answering track report.
\newblock In \emph{In Proceedings of TREC-8}, pages 77--82.

\bibitem[{Wang et~al.(2021)Wang, Thakur, Reimers, and Gurevych}]{wang2021gpl}
Kexin Wang, Nandan Thakur, Nils Reimers, and Iryna Gurevych. 2021.
\newblock \href {http://arxiv.org/abs/2112.07577} {{GPL:} generative pseudo
  labeling for unsupervised domain adaptation of dense retrieval}.
\newblock \emph{CoRR}, abs/2112.07577.

\bibitem[{Wang et~al.(2007)Wang, Smith, and Mitamura}]{wang2007jeopardy}
Mengqiu Wang, Noah~A Smith, and Teruko Mitamura. 2007.
\newblock What is the jeopardy model? a quasi-synchronous grammar for qa.
\newblock In \emph{Proceedings of the 2007 Joint Conference on Empirical
  Methods in Natural Language Processing and Computational Natural Language
  Learning (EMNLP-CoNLL)}, pages 22--32.

\bibitem[{Wang and Jiang(2017)}]{Wang2017ACM}
Shuohang Wang and Jing Jiang. 2017.
\newblock \href {https://openreview.net/forum?id=HJTzHtqee} {A
  compare-aggregate model for matching text sequences}.
\newblock In \emph{5th International Conference on Learning Representations,
  {ICLR} 2017, Toulon, France, April 24-26, 2017, Conference Track
  Proceedings}. OpenReview.net.

\bibitem[{Xu et~al.(2021)Xu, Liang, Huang, and
  Xiang}]{DBLP:journals/corr/abs-2110-06393}
Peng Xu, Davis Liang, Zhiheng Huang, and Bing Xiang. 2021.
\newblock \href {http://arxiv.org/abs/2110.06393} {Attention-guided generative
  models for extractive question answering}.
\newblock \emph{CoRR}, abs/2110.06393.

\bibitem[{Yang et~al.(2015)Yang, Yih, and Meek}]{yang2015wikiqa}
Yi~Yang, Wen-tau Yih, and Christopher Meek. 2015.
\newblock Wikiqa: A challenge dataset for open-domain question answering.
\newblock In \emph{Proceedings of the 2015 conference on empirical methods in
  natural language processing}, pages 2013--2018.

\bibitem[{Yoon et~al.(2019)Yoon, Dernoncourt, Kim, Bui, and Jung}]{Yoon2019ACM}
Seunghyun Yoon, Franck Dernoncourt, Doo~Soon Kim, Trung Bui, and Kyomin Jung.
  2019.
\newblock A compare-aggregate model with latent clustering for answer
  selection.
\newblock \emph{Proceedings of the 28th ACM International Conference on
  Information and Knowledge Management}.

\bibitem[{Zhang* et~al.(2020)Zhang*, Kishore*, Wu*, Weinberger, and
  Artzi}]{Zhang*2020BERTScore:}
Tianyi Zhang*, Varsha Kishore*, Felix Wu*, Kilian~Q. Weinberger, and Yoav
  Artzi. 2020.
\newblock \href {https://openreview.net/forum?id=SkeHuCVFDr} {Bertscore:
  Evaluating text generation with bert}.
\newblock In \emph{International Conference on Learning Representations}.

\bibitem[{Zhang et~al.(2021)Zhang, Vu, and Moschitti}]{zhang-etal-2021-joint}
Zeyu Zhang, Thuy Vu, and Alessandro Moschitti. 2021.
\newblock \href {https://doi.org/10.18653/v1/2021.acl-long.252} {Joint models
  for answer verification in question answering systems}.
\newblock In \emph{Proceedings of the 59th Annual Meeting of the Association
  for Computational Linguistics and the 11th International Joint Conference on
  Natural Language Processing (Volume 1: Long Papers)}, pages 3252--3262,
  Online. Association for Computational Linguistics.

\bibitem[{Zhang et~al.(2022)Zhang, Vu, and Moschitti}]{trecqazeyu}
Zeyu Zhang, Thuy Vu, and Alessandro Moschitti. 2022.
\newblock \href {http://arxiv.org/abs/2201.05981} {Double retrieval and ranking
  for accurate question answering}.
\newblock \emph{CoRR}, abs/2201.05981.

\end{thebibliography}
\bibliographystyle{acl_natbib}

\clearpage
\appendix
\section*{Appendix}

\section{Experimental setup details}
\label{apx:exp_setup}

\subsection{Hyperparameter Selection}
\label{ref:apx_hyperparameter_selection}
For our experiments, we consider different combinations of hyper-parameters. In particular, for training we combined the following parameters in multiple experiments: $lr \in \{1e^{-6}, 5e^{-5}, 1e^{-4}\}$, $k \in \{5, max\}$, $\text{batch size} \in \{64, 128, 256\}$, $\text{precision} \in \{16, 32\}$ along with using both Adam \cite{Kingma2015AdamAM} and Adafactor \cite{Shazeer2018AdafactorAL} optimizers. We selected the best combination of hyper-parameters (described in Section~\ref{sec:experiments}) by manually evaluating the output of different models on a reduced version of the MS-MARCO NLG dataset. Additionally, we also experimented with different hyper-parameters for the decoder. Specifically, we performed a qualitative evaluation of the answers generated using different parameters to select the best configuration: beam search of $5$, and forcing the generated answer to have a number of tokens in the interval $[6,100]$.

\subsection{Dataset Statistics}
\label{ref:apx_data_statistics}
Below we provide the statistics for the different datasets that we use in our experiments. 
\begin{table}[ht]
\tiny
\centering
\resizebox{1\linewidth}{!}{
\begin{tabular}{lcccc}
\toprule
Dataset       &   Split    & \# of Q & QA Pairs \\
\toprule
MS-MARCO QA      & Train           & 655006   & 19543964 \\
MS-MARCO QA      & Dev             & 1000    & 29309  \\
\midrule
MS-MARCO NLG    & Train            & 153725   & 4694170 \\
MS-MARCO NLG    & Dev              & 1000     & 28353 \\
MS-MARCO NLG    & Test             & 1000     & 28529 \\
\midrule
WikiQA        & Train (clean)      &  857        & 8651  \\
WikiQA        & Dev (clean)        &  121        & 1126 \\
WikiQA        & Test               &  633        & 6165 \\
\midrule
TrecQA        & Train              & 804      & 32964 \\
TrecQA        & Dev                & 216      & 9590  \\
TrecQA        & Test               & 340      & 13416 \\
\bottomrule
\end{tabular}}
\caption{\small Datasets statistics.}
\label{tab:datasets_statistics}
\end{table}

\subsection{Computational Setup}
We perform each experiment on $8$ NVIDIA A100 GPUS (40GB RAM) using DDP\footnote{\url{https://pytorch.org/docs/master/generated/torch.\\nn.parallel.DistributedDataParallel.html}} as distributed training strategy. To complete the training on MS-MARCO every model takes approximately 6 days, while the experiments on TREC-QA and WikiQA takes 4 hours.

\section{Checkpoint Selection using AS2 Model}
\label{apx:ckpt_selection}

\begin{figure}[ht!]
    \begin{center}
        \begin{tikzpicture}
        \begin{axis}[width=0.9\linewidth, xlabel=Steps, ylabel=loss (validation)]
            \addplot table [x=steps, y=devloss, col sep=comma] {csv/devloss_lwsilfsolfv3.csv};
            \addplot +[mark=none,style=dashed] coordinates {(22500, 1.22) (22500, 1.28)};
        \end{axis}
        \label{plt:devloss_ckpt}
        \end{tikzpicture}
        \begin{tikzpicture}
        \begin{axis}[width=0.9\linewidth, xlabel=Steps, ylabel=$AVG(AS2_{Score})$]
            \addplot table [x=steps, y=rrmean, col sep=comma] {csv/rr_lwsilfsolfv3.csv};
            \addplot +[mark=none,style=dashed] coordinates {(60000, 0.78) (60000, 1)};
        \end{axis}
        \label{plt:rr_ckpt}
        \end{tikzpicture}
    \caption{\small These plots show the differences between using the dev. loss and the average AS2 model score as the measure for model checkpoint selection. Notice that the vertical dashed red line is used to identify the best checkpoint (which is the one with the lowest loss in the first plot, and the one with the highest average AS2 score in the second).}
    \label{plt:checkpointselection}
    \end{center}
\end{figure}
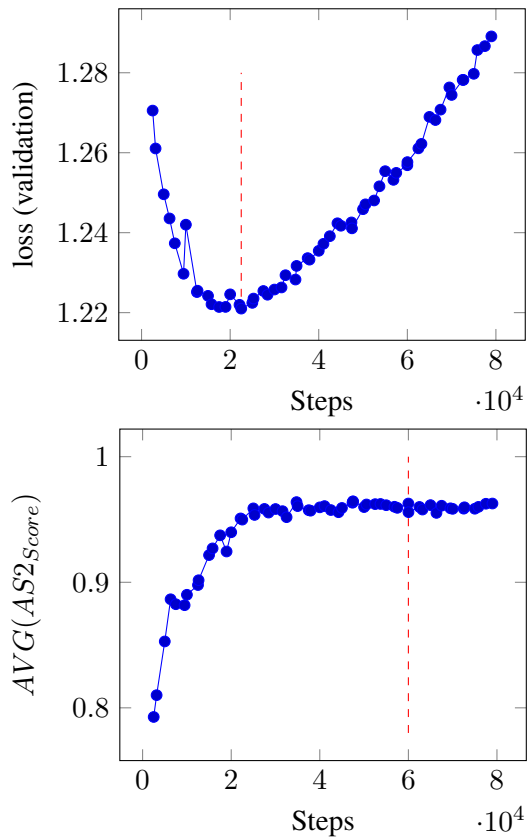

We observed that minimizing the loss on the development split does not strictly correlate with better answer generation from a human annotation perspective. Thus we experimented with an alternate performance measure for model checkpoint selection. Specifically, we use each checkpoint to generate outputs for a number of validation examples and score them with an AS2 model. We use the average scores produced by the AS2 model as the metric for deciding the best model checkpoint.  The differences between using the loss on the development split and the average AS2 score can be noticed in Figure \ref{plt:checkpointselection}. We conducted a manual evaluation of outputs for different checkpoints and determined that using AS2 scores correlates better with our manual judgements than development set loss. We plan to explore this technique further in future work.

\section{Details of Human Evaluation}
\label{apx:maneval}
To evaluate our models we used the Amazon Mechanical Turk\footnote{\url{https://www.mturk.com/}} framework. We designed an annotation task in which we showed to a pool of several high quality annotators (turkers) a question, a target answer generated from our models and, where possible (e.g., MS MARCO-NLG), a well-formed reference answer asking if the target answer was correct or not. For each QA pair (hit) we paid $0.1\$$ and we assigned $5$ turkers. Specifically, we selected each Turker considering only masters with an approval rate greater than $95\%$ and with at least $500$ hits approved.

\section{Human Annotations v/s BLEU}
\label{apx:acc_vs_bleu}

Comparing human evaluation to BLEU scores for GenQA model outputs, we find that BLEU is not a reliable performance metric for this task and setting. For BLEU, we use two references for the generated target answer: (i) the manually written answer from MS-MARCO NLG, and (ii) the top ranked answer by the AS2 model that is being used as the teacher for knowledge transfer. The results are shown in Table \ref{tab:bleu_target} and appear to be quite random. Neither of the rankings induced by the BLEU metric correspond to the ranking induced by human evaluation (see Table~\ref{tab:msmarco_results}).

{\color{red}
\begin{table}
\begin{center}
    \resizebox{1\linewidth}{!}{
    \begin{tabular}{lcccc} 
        \toprule
        Model  & Transfer & Supervised & BLEU       & BLEU \\
           &          &            & NLG answer & AS2 answer \\
        \toprule
        %       &         &          &          & 100.00 & 30.97 \\
        RoBERTa &  -       & -          & 34.76  & 100.0 \\
        \midrule
        T5       & {\WS}       &  -       & 46.67  & 32.69 \\
        T5       & {\WS}+{\LW}    &  -       & 38.47  & 65.55  \\
        T5       & {\WS}+{\LW}+{\SC} &  -        & 38.85  & 63.76  \\
        \midrule
        T5       &   -       & \cite{hsu2021answer} & 36.94  & 61.22  \\
        T5       & {\WS}+{\LW}+{\SC} & \cite{hsu2021answer}  & 49.33  & 37.62  \\
        \bottomrule
    \end{tabular}
    }
    \caption{\small Results using BLEU as the metric for measuring performance of answer generation. We train the GenQA models on MS-MARCO NLG for the supervised setting, and MS-MARCO QA for the knowledge transfer setting.}
    \label{tab:bleu_target}
\end{center}
\end{table}
}

\section{Qualitative Evaluation}
\label{apx:quality_eval}

In this section, we present some qualitative examples of answers generated by our models. In particular, we show (i) the differences between the generated answer with the answer candidates used as input for GenQA, and (ii) how we can manipulate the generated answers by forcing the decoding to start from different {\SCO} bucket label tokens.

\subsection{Generated Answer v/s Input Candidates}

From qualitative analysis we observe that the answers generated by our knowledge transfer techniques are generally longer than the answers generated by a GenQA model trained in a fully supervised manner. We present three examples in Table~\ref{tab:generated_answers}.  In the first example, both the GenQA answers are correct, and we observe that the AS2 selected top answer contains the correct answer string in it. In the second example, both the GenQA answers are incorrect, however, similar to the previous example, both the generated answers are shorter and syntactically improved versions of the AS2 selected answer. In the third example, the answer generated by our weakly supervised GenQA model is correct (the AS2 selected top answer is also correct) while that generated by the supervised GenQA model is incorrect. Overall, we notice that our weakly supervised models tend to copy and summarize the answer from the input candidates having the highest AS2 scores, while the supervised GenQA model generates more concise and shorter answers.

\subsection{Forced Decoding using {\SCO}}

In this section, we aim to analyze how the {\SCO} bucket tokens can be used to  modify the quality of the generated answers from GenQA models. Specifically, we tested our GenQA models trained with {\SCO} by forcing the decoder to generate answers starting from each of the different {\SCO} bucket tokens: (\{{\YES}, {\PROBABLY}, {\MAYBE}, {\DOUBT}, {\NO}\}). We present an anecdotal example in Table~\ref{tab:force_answers}. We observe that the syntactical quality of the generated answers correlates with the {\SCO} bucket token selected as the first token of the generated answer. Furthermore, higher confidence {\SCO} bucket tokens (e.g., {\YES}, {\PROBABLY}) tend to generate shorter and more concise answers, while lower confidence {\SCO} bucket tokens like {\DOUBT} and {\NO} can be used to generate longer sequences that are syntactically inferior.

\begin{table*}[ht]
\small
\begin{center}
\resizebox{0.9 \linewidth}{!}{
    \begin{tabular}{rl} 
        \toprule
        \multicolumn{2}{c}{\textbf{"what city had a world fair in 1900"}} \\
        $[\_\text{YES}\_]$      & "The 1900 world's fair was held in Paris, France." \\
        $[\_\text{PROBABLY}\_]$ & "The 1900 world fair was held in Paris, France." \\
        $[\_\text{MAYBE}\_]$    & "the 1900 world's fair, in paris which opened on march 19, 1900 is still one of the ... \\
                         & most important and influential exhibitions in modern history." \\
        $[\_\text{DOUBT}\_]$    & "in 1902 the exhibition denoting the new century at paris had been reopened to the public ... \\
                         & and there were only 12 sculptures by Louis britton and eight paintings by marcus thomas." \\
        $[\_\text{NO}\_]$       & "after a year of years' recovery but still having trouble finding an adequate buyer, the glass ...\\
                         & was shipped to france in 1900 for the world fair in paris and will be forever lost on this museum trail." \\
       \bottomrule
    \end{tabular}}
    \caption{\small An anecdotal example of GenQA using {\SCO} by forcing the decoder to start the answer generation from different {\SCO} bucket tokens. We observe that the high confidence {\SCO} bucket answer is concise and correctly answers the question without the need of any further reasoning.}
    \label{tab:force_answers}
\end{center}
\end{table*}
\begin{table*}[t]
\small
\begin{center}
\resizebox{0.9 \linewidth}{!}{
    \begin{tabular}{ll} 
        \toprule
        \multicolumn{2}{c}{\textbf{Inputs}} \\
        \textbf{Question:}   & when was tom sawyer written \\
        \textbf{C1:} (0.99)  & Among his novels are The Adventures of Tom Sawyer (1876) and its sequel, Adventures... \\
                             & of Huckleberry Finn (1885), the latter often called The Great American Novel. \\
        \textbf{C2:} (0.71)  & When it first came out in 1876, however, it was comparatively a failure. \\
        \textbf{C3:} (0.34)  & Tom Sawyer (1973) G | 1h 43min | Adventure, Musical, Family | 15 March 1973 (USA) Tom Sawyer and... \\
                             & his pal Huckleberry Finn have great adventures on the Mississippi River, pretending to be pirates,... \\
                             & attending their own funeral, and witnessing a murder. \\
        \textbf{C4:} (0.06)  & When he was four, Twain's family moved to Hannibal, Missouri, a port town on the Mississippi River...\\
                             & that inspired the fictional town of St. Petersburg in The Adventures of Tom Sawyer and Adventures...\\
                             & of Huckleberry Finn. \\
        \textbf{Gold answer:} & Tom sawyer written is on 1876.\\
        \multicolumn{2}{c}{\textbf{Generated \& Selected}} \\
        \textbf{AS2 Selection:}  & Among his novels are The Adventures of Tom Sawyer (1876) and its sequel, Adventures... \\
                                        & of Huckleberry Finn (1885), the latter often called The Great American Novel. \\
        \textbf{GenQA~\cite{hsu2021answer}:}                 & Tom Sawyer was written in 1876. \\
        \textbf{(Ours) WS+LW+SC:}              & The Adventures of Tom Sawyer is a novel by Mark Twain, first published in 1876. \\
        \\
        \toprule
        \multicolumn{2}{c}{\textbf{Inputs}} \\
        \textbf{Question:}   & which planet is colder saturn or neptune \\
        \textbf{C1:} (0.98)  & Saturn: Due to its distance from the Sun, Saturn is a rather cold gas giant ...\\
                             & planet, with an average temperature of -178 °Celsius \\
        \textbf{C2:} (0.90)  & Neptune: With temperatures dropping to -218°C in Neptune’s upper atmosphere, ...\\
                             & the planet is one of the coldest in our Solar System. \\
        \textbf{C3:} (0.79)  & The next four planets we will talk about--Jupiter, Saturn, Uranus, and Neptune--are ... \\
                             & very different to the terrestrial planets in that they have no real solid surfaces,... \\
                             & being dominated by thick, deep atmospheres. \\
        \textbf{C4:} (0.74)  & Jupiter is the fifth planet from the Sun, and is the largest planet in the solar system. \\
        \textbf{Gold answer:} & The planet Saturn is colder. \\
        \multicolumn{2}{c}{\textbf{Generated \& Selected}} \\
        \textbf{AS2 Selection:}  & Saturn: Due to its distance from the Sun, Saturn is a rather cold gas giant ...\\
                                 & planet, with an average temperature of -178 °Celsius \\
        \textbf{GenQA~\cite{hsu2021answer}:}          & Neptune is colder than Saturn. \\
        \textbf{(Ours) WS+LW+SC:}       & Uranus is the fifth planet from the Sun, and is the largest planet in the solar system. \\
        \\
        \toprule
        \multicolumn{2}{c}{\textbf{Inputs}} \\
        \textbf{Question:}   & average of rn \\
        \textbf{C1:} (0.99)  & According to the Bureau of Labor Statistics lastest data, the average salary of a registered ... \\
                             & nurse in the United States is \$69,790. \\
        \textbf{C2:} (0.98)  & The starting salary of a Registered Nurse can range from around \$28,000-\$50,000 per year on average. \\
        \textbf{C3:} (0.90)  & The starting hourly wage of an RN can range from \$16.50-26.00 per hour. \\
        \textbf{C4:} (0.81)  & Although the average annual salary of a registered nurse in 2011 was \$69,110, according to the ... \\
                             & BLS, salaries are affected by a number of factors. \\
        \textbf{Gold answer:} & The average salary of a registered nurse is \$69,790. \\
        \multicolumn{2}{c}{\textbf{Generated \& Selected}} \\
        \textbf{AS2 Selection:}  & According to the Bureau of Labor Statistics lastest data, the average salary of a registered ... \\
                                        & nurse in the United States is \$69,790. \\
        \textbf{GenQA~\cite{hsu2021answer}:}                 & The average salary of a registered nurse is \$28,000 to \$50,000 per year. \\
        \textbf{(Ours) WS+LW+SC:}              & The average salary of a Registered Nurse is \$69,790 per year. \\
        \bottomrule
    \end{tabular}}
    \caption{\small Anecdotal examples from MS-MARGO NLG test set with answers generated from (i) our weakly supervised GenQA model (trained using  WS+LW+SC), (ii) a fully supervised GenQA model~\cite{hsu2021answer}, (iii) the AS2 selected top answer and (iv) the manually written gold answer in the data set. We indicate the AS2 model scores for the input answer candidates to better reason about these examples.}
    \label{tab:generated_answers}
\end{center}
\end{table*}

\end{document}